\documentclass[twoside]{article}
\usepackage[accepted]{style/aistats2022}
\usepackage[american]{babel}

\usepackage{amsmath,amsfonts,bm}

\def\eqref#1{equation~\ref{#1}}

\def\1{\bm{1}}

\def\bX{{\textbf{X}}}
\def\bY{{\textbf{Y}}}
\def\bU{{\textbf{U}}}
\def\bF{{\textbf{F}}}

\DeclareMathAlphabet{\mathsfit}{\encodingdefault}{\sfdefault}{m}{sl}
\SetMathAlphabet{\mathsfit}{bold}{\encodingdefault}{\sfdefault}{bx}{n}

\def\gN{{\mathcal{N}}}

\newcommand{\kl}{{\mathrm{KL}}}

\usepackage[square]{natbib}
    \renewcommand{\bibsection}{\subsubsection*{References}}
\usepackage{mathtools} 
\usepackage{microtype}
\usepackage{graphicx}
\usepackage[utf8]{inputenc} 
\usepackage[T1]{fontenc}    
\usepackage{tabularx}
\usepackage{comment}
\usepackage{mathtools}
\usepackage{cancel}
\usepackage{graphicx}
\usepackage{array,booktabs}
\usepackage{amsmath}
\usepackage{amssymb,amsfonts}
\usepackage{standalone}
\usepackage{xcolor}
\usepackage{multirow}
\usepackage{array}
\DeclareMathAlphabet{\mathcal}{OMS}{cmsy}{m}{n}
\usepackage[ruled,vlined ]{algorithm2e}
\usepackage{tikz}
\usetikzlibrary{bayesnet}
\usepackage{float}
\usepackage{amssymb}
\usepackage{pifont}
\newcommand{\xmark}{\text{\ding{55}}}
\newcommand{\cmark}{\text{\ding{51}}}
\newcommand{\red}[1]{\textcolor{red}{#1}}
\usepackage{subfigure}
\usepackage{booktabs} 

\usepackage[belowskip=-10pt,aboveskip=4pt]{caption}
\makeatletter
\newcommand{\longdash}[1][2em]{%
  \makebox[#1]{$\m@th\smash-\mkern-7mu\cleaders\hbox{$\mkern-2mu\smash-\mkern-2mu$}\hfill\mkern-7mu\smash-$}}
\makeatother
\newcommand{\omitskip}{\kern-\arraycolsep}

\newcommand{\rlongdash}[1][2em]{\omitskip\longdash[#1]}
\global\hbadness=10000
\usepackage[colorlinks=true,citecolor=black, linkcolor=black]{hyperref}%
\usepackage{cleveref}

\begin{document}

\twocolumn[

\aistatstitle{Generalised Gaussian Process Latent Variable Models (GPLVM) with Stochastic Variational Inference}

\aistatsauthor{Vidhi Lalchand \And Aditya Ravuri \And  Neil D. Lawrence}

\aistatsaddress{University of Cambridge \And University of Cambridge \And University of Cambridge}]

\begin{abstract}

Gaussian process latent variable models (GPLVM) are a flexible and non-linear approach to dimensionality reduction, extending classical Gaussian processes to an unsupervised learning context. The Bayesian incarnation of the GPLVM \citep{titsias2010bayesian} uses a variational framework, where the posterior over latent variables is approximated by a well-behaved variational family, a factorised Gaussian  yielding a tractable lower bound. However, the non-factorisability of the lower bound prevents truly scalable inference. In this work, we study the doubly stochastic formulation of the Bayesian GPLVM model amenable with minibatch training. We show how this framework is compatible with different latent variable formulations and perform experiments to compare a suite of models. Further, we demonstrate how we can train in the presence of massively missing data and obtain high-fidelity reconstructions.We demonstrate the model's performance by benchmarking against the canonical sparse GPLVM for high dimensional data examples. 

\end{abstract}

\section{Introduction}

Gaussian processes (GPs) represent a powerful non-parametric probabilistic framework for performing regression and classification. The inductive biases are controlled by a kernel function \citep{rasmussen2004gaussian}. The Gaussian process latent variable model (GPLVM) \citep{lawrence2004gaussian} paved the way for GPs to be used in unsupervised learning tasks like dimensionality reduction and structure discovery for high-dimensional data. It provides a probabilistic mapping from (an unobserved) latent space ($\bX$) to data-space ($\bY$). The GP acts as a \textit{decoder} and the smoothness of the mapping is controlled by a kernel function. Many traditional dimensionality reduction models learn a projection of high dimensional data to lower dimensional manifolds. In the GPLVM the direction of the mapping is reversed. 

The standard GPLVM is a multi-output regression model where the inputs are unobserved during training. The canonical formulation treats the unknown latent variables as point estimates and optimizes the marginal likelihood jointly with the covariance hyperparameters ($\bm{\theta}$). Techniques to apply Gaussian processes to very large datasets were introduced in \citet{hensman2013gaussian} which demonstrated how stochastic variational inference (SVI) \citep{JMLR:v14:hoffman13a} can be used with sparse GPs in a regression context. The key idea is to re-formulate the evidence lower bound (ELBO) in a way that factorizes across the data enabling mini-batching for gradients. The canonical formulation can be made sparse by using the regression based lower bound from \citet{hensman2013gaussian} and optimising for latents $\bX$. We call this model the \textit{Sparse GPLVM} or \textsc{Point} for short. We also study the performance of maximum-a-posteriori (MAP) in this framework.

The Bayesian formulation of the GPLVM in \citep{titsias2010bayesian} variationally integrates out latent variables, providing principled uncertainty around the latent encoding. This formulation relies on inducing variables \citep{titsias2009variational} that admit a tractable lower bound while providing computational savings. The Bayesian formulation also allows the dimensionality of the latent space to be automatically determined by using the standard \textit{automatic relevance determination} squared exponential (SE-ARD) kernel whose lengthscales are determined by maximisation of the ELBO. Extraneous dimensions acquire longer lengthscales and are automatically pruned. However, this closed form framework does not factorise across data points \citep{titsias2010bayesian} preventing the application of Bayesian GPLVM to larger datasets.

In this paper we extend the big data regression setting proposed in \citet{hensman2013gaussian} to the unsupervised latent variable model setting. We re-formulate Bayesian GPLVM for scalable inference using SVI by using a structured doubly stochastic lower bound \citep{titsias2014doubly}. We denote this model as \textit{Bayesian SVI} or \textsc{B-SVI} for short.

The smooth GP decoder mapping ensures that points close in latent space are mapped to points close in data space. The notion of an \textit{encoder} for GPLVMs was introduced in \citep{lawrence2006local} where an additional mapping (called the \textit{back-constraint} by the authors) was learnt expressing each latent point in the evidence (marginal likelihood) as a function of its corresponding data point. This incarnation ensured that data-space proximities were preserved in latent encodings. Hence, GPLVMs can be put on the same footing as autoencoding models with an \textit{encoder} mapping from data to latent space and a \textit{decoder} mapping from latent to data space. Such a model was considered in \citet{bui2015stochastic} and this is the fourth model we include in our compendium which we call \textit{Autoencoded Bayesian SVI} or \textsc{AEB-SVI}.
In summary, our main contributions are:
\vspace{-2mm}
\begin{itemize}
\itemsep0em 
    \item Present a generalised framework for GPLVM models which differ in the form of the latent variable set-up and but share the same inference strategy (SVI). We conduct experiments with the SVI-compatible doubly stochastic evidence lower bound for the \textsc{Point}, maximum-a-posteriori \textsc{(MAP)}, Bayesian SVI (\textsc{B-SVI}) and \textsc{AEB-SVI} models enabling efficient and scalable inference. 
    \item Extend this framework to dimensionality reduction for non-conjugate likelihoods across all the latent variable incarnations.  
    \item Demonstrate how training in these models is compatible with partially and massively missing data settings\footnote{bulk of the dimensions missing for every data point yielding a very sparse data matrix.} frequently embodied in real-world datasets.
\end{itemize}
\vspace{-3mm}
\begin{table*}[t]  
   \caption{\small{Existing approaches for Inference in GPLVMs. Our work studies the scalable alternative with SVI across all these models. The decoder $(X \longrightarrow Y)$ is a GP across all methods.}}
    \label{pastwork}
    \centering
    \resizebox{0.95\textwidth}{!}{
    \begin{tabular}{c|c|c|c|c}
          Reference  & Data Likelihood & Latent Variable $q(X)$ & Encoder ($Y \rightarrow X$) & Training Method  \\
         \hline 
         \citet{lawrence2004gaussian} & Gaussian & point est. & \xmark & Gradient descent\\
        \citet{lawrence2006local}  &  Gaussian & point est.  & $\cmark$ & Gradient descent\\
         \citet{titsias2010bayesian} & Gaussian & Gaussian  & \xmark & Collapsed VI \\
         \citet{bui2015stochastic} &  Gaussian & Gaussian  & $\cmark$ & SVI\\ 
         \citet{ramchandran2021latent} & Any & Gaussian  & $\cmark$ & SVI\\ 
       \textbf{This work} & Any & point / Gaussian & $\xmark$/$\cmark$ & SVI
    \end{tabular}}
    \vspace{-5mm}
 \end{table*}
\section{Background}
\subsection{Bayesian GPLVM}
\label{bgpl}
In the sparse variational formulation underlying the Bayesian GPLVM we have a training set comprising of $N$ $D$-dimensional real valued observations $\bY \equiv \{\bm{y}_{n}\}_{n=1}^{N} \in \mathbb{R}^{N \times D}$. These data are associated with $N$ $Q$-dimensional latent variables, $\bX \equiv \{\bm{x}_{n}\}_{n=1}^{N}\in \mathbb{R}^{N \times Q}$ where $Q < D$ provides dimensionality reduction \citep{lawrence2004gaussian}. The forward mapping ($\bX \longrightarrow \bY$) is governed by GPs independently defined across dimensions $D$. The sparse GP formulation describing the data is as follows:
\scalebox{0.99}{
\begin{minipage}{\linewidth}
\begin{align}
p(\bX) &= \displaystyle \prod _{n=1}^N \gN (\bm{x}_{n};\bm{0}, \mathbb{I}_{Q}), \nonumber\\
p(\bF|\bU, \bX, \bm{\theta}) &= \displaystyle \prod_{d=1}^{D}\mathcal{N}(\bm{f}_{d}; K_{nm}K_{mm}^{-1}\bm{u}_{d}, Q_{nn}),\\
p(\bY| \bF, X) &= \prod_{n=1}^N \prod_{d=1}^D \mathcal{N}(y_{n,d}; \bm{f}_{d}(\bm{x}_{n}), \sigma^{2}_{y}),\nonumber
\label{initial}
\end{align}
\end{minipage}}

where $Q_{nn} = K_{nn} - K_{nm}K_{mm}^{-1}K_{mn}$, $\bF \equiv \{ \bm{f}_{d} \}_{d=1}^{D}$, $\bU \equiv \{\bm{u}_{d} \}_{d=1}^{D}$ and $\bm{y_{d}}$ is the $d^{th}$ column of $\textbf{Y}$. $K_{nn}$ is the covariance matrix corresponding to a user chosen positive-definite kernel function $k_{\theta}(x, x^{\prime})$ evaluated on latent points $\{\bm{x}_{n}\}_{n=1}^{N}$ and parameterised by hyperparameters $\bm{\theta}$. The kernel hyperparameters are shared across all dimensions $D$. 

The inducing variables per dimension $\{ \bm{u}_{d} \}_{d=1}^{D}$ are distributed with a GP prior $\bm{u}_{d}|Z \sim \mathcal{N}(\bm{0}, K_{mm})$ computed on inducing input locations  $Z \in \mathbb{R}^{M \times Q}$ which live in latent space and have dimensionality $Q$ (matching $\bm{x}_{n}$).
The variational formulation,
\begin{equation}
\begin{split}
 p(\bF, \bX, \bU |\bY) &= \Big [\prod_{d=1}^{D}p(\bm{f}_{d}|\bm{u}_{d},X)q(\bm{u}_{d}) \Big]q(\bX)  \hspace{-4mm} \\
 & \approx q(\bF, \bX, \bU) 
\label{titsias}
\end{split}
\end{equation}
admits a tractable lower bound to the marginal likelihood $p(\bY|\bm{\theta})$ where the inducing variables are integrated out or \textit{collapsed} \citep{titsias2010bayesian}. 
\begin{figure}
    \includegraphics[scale=0.32]{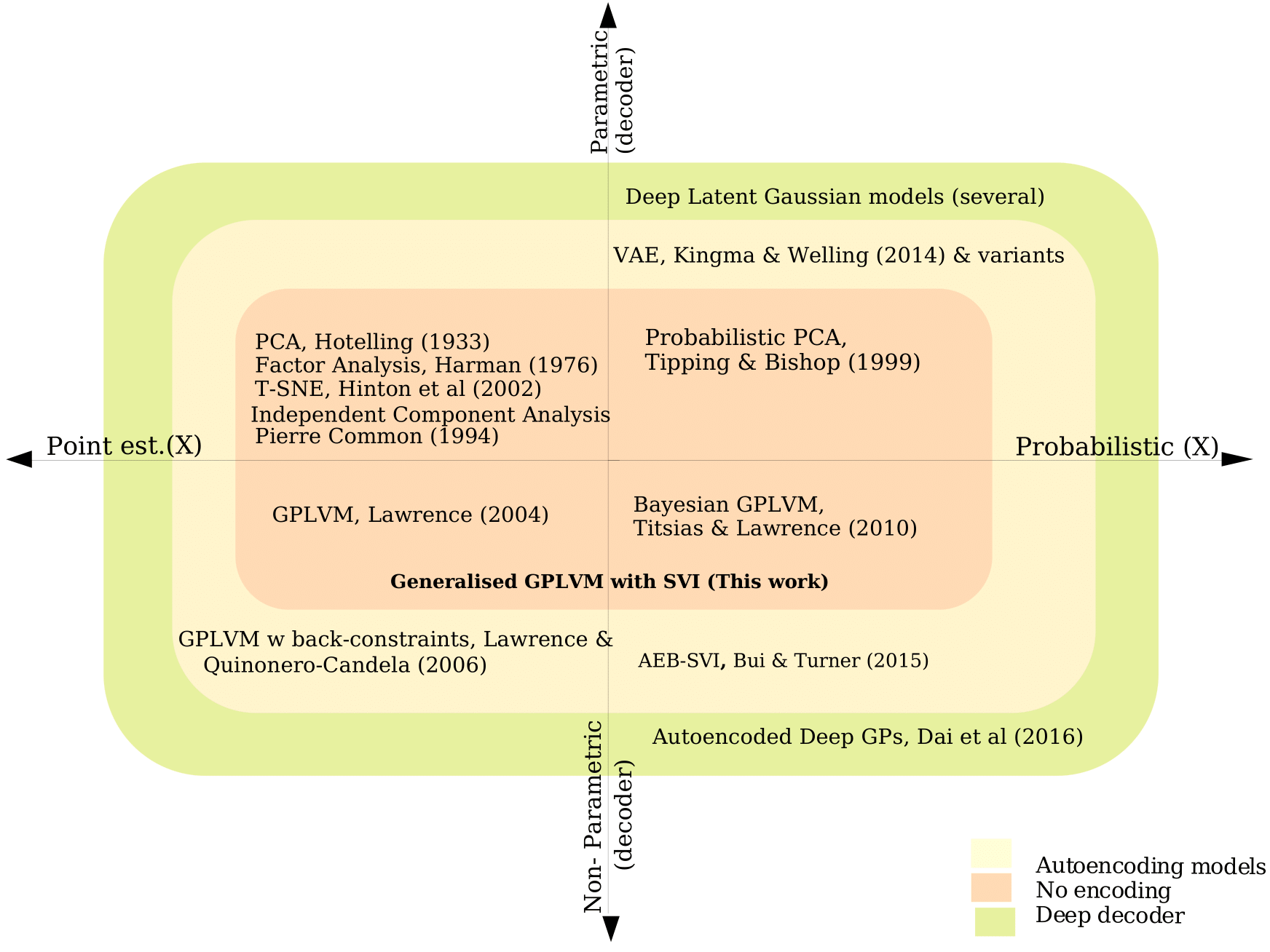}
    \caption{A taxonomy of latent variable models for unsupervised dimensionality reduction along three axis of variation. 1) the form of the latent variable, 2) the nature of the decoder and 3) whether or not the models are autoencoding. The framework in this work is amenable with point estimation and Bayesian learning as well as amortisation.}
    \label{fig:my_label}
\end{figure}
The original bound incorporated the optimal Gaussian variational distribution $q^{*}(\bm{u}_{d})$ and a diagonal Gaussian variational distribution, $q(\bX) =\prod_{n=1}^{N}\mathcal{N}(\bm{x}_{n}; \mu_{n}, s_{n}\mathbb{I}_{Q}), $. However, every gradient step needs a pass over the full dataset of size $N$. In the section below we describe the Bayesian SVI model which uses the same variational formulation as above except for the treatment of the inducing variables per dimension $\bm{u}_{d}$. Instead of using their optimal analytic form, we learn their parameters through direct optimisation of the \textit{uncollapsed} lower bound. Beyond speeding up inference, the uncollapsed bound has properties which open up several possibilities, for instance, training with high-dimensional data with a non-Gaussian likelihood and structure discovery in the presence of sparse, high-dimensional data.

\section{Generalised GPLVM with SVI}

The key insight from \citet{hensman2013gaussian} is to keep the representation of $\bU$ uncollapsed and learn $q(\bm{u}_{d}) \sim \mathcal{N}(\bm{m}_{d}, S_{d})$ numerically using stochastic gradient methods. In the next sections, we extend this insight to variationally learning $q(\textbf{X})$.
\subsection{Is SVI applicable?}
Stochastic Variational Inference (SVI) \citep{JMLR:v14:hoffman13a} pre-requisites a joint probability model with a set of global and local hidden variables where the local variables are conditionally independent given the global variables. GP models for regression in their standard form do not admit such a factorisation and neither do they have global variables, however \citet{hensman2013gaussian} showed how the SVI machinery becomes applicable by introducing global inducing variables $\bm{u}$ and variationally marginalising $\bm{f}$. We assume a single output dimension in this sub-section for clarity, hence drop the dimension index $d$.
\vspace{-2mm}
\begin{align}
    \ln p(\bm{y}|\bm{u}) &= \ln \int p(\bm{y}|\bm{f})p(\bm{f}|\bm{u})d\bm{f}
    \geq \mathbb{E}_{p(\bm{f}|\bm{u})}[\ln p(\bm{y}|\bm{f})] \nonumber \\
    &\triangleq \ln \tilde{p}(\bm{y}|\bm{u})  = \prod_{n=1}^{N}\mathcal{N}(y_{n}|k_{n}^{T}K_{mm}^{-1}\bm{u}, \sigma^{2}_{y})\times \nonumber \\ &\exp\left\{-\dfrac{1}{2\sigma^{2}_{y}}(k_{nn} - k_{n}^{T}K_{mm}^{-1}k_{n}) \right\} 
\end{align}
where $\tilde{p}(\bm{y}|\bm{u})$ factorises if the likelihood $p(\bm{y}|\bm{f})$ does and $k_{n}$ is the $n^{th}$ column of $K_{mn}$ (only dependent on point $\bm{x}_{n}$). We now have a model with global variables and a likelihood which is conditionally independent across observations given the global variables $\bm{u}$. The regression model does not need local hidden variables. However, in the latent variable setting we have a latent variable $\bm{x}_{n}$ per training point. \vspace{-3mm}
\subsection{Doubly Stochastic Evidence Lower bound (DS-ELBO)}
\label{rudimentary}

The term \textit{doubly stochastic inference} was proposed by \citet{titsias2014doubly} and deployed in deep Gaussian process regression by \citet{salimbeni2017doubly}. Here we use doubly stochastic inference in the unsupervised latent variable setting, where the goal is dimensionality reduction.

Keeping with the formulation in section \ref{bgpl} we write down the rudimentary ELBO, with the familiar decomposition involving the expected log-likelihood term and KL terms,

\scalebox{0.85}{
\begin{minipage}{\linewidth}
\begin{align}\label{elbo}
\mathcal{L} &= \int p(\bF|\bU, \bX)q(\bU)q(\bX) \log \dfrac{p(\bY|\bF, \bX)p(\bU|Z)p(\bX)}{q(\bU)q(\bX)}d\bF d\bU d\bX \\
 &= \mathbb{E}_{q(.)}[\log p(\bY|\bF, \bX)] - \kl(q(\bX)||p(\bX)) - \kl(q(\bU)||p(\bU)) \nonumber
\end{align}
\end{minipage}}
where $q(.)$ is as in \cref{titsias}.
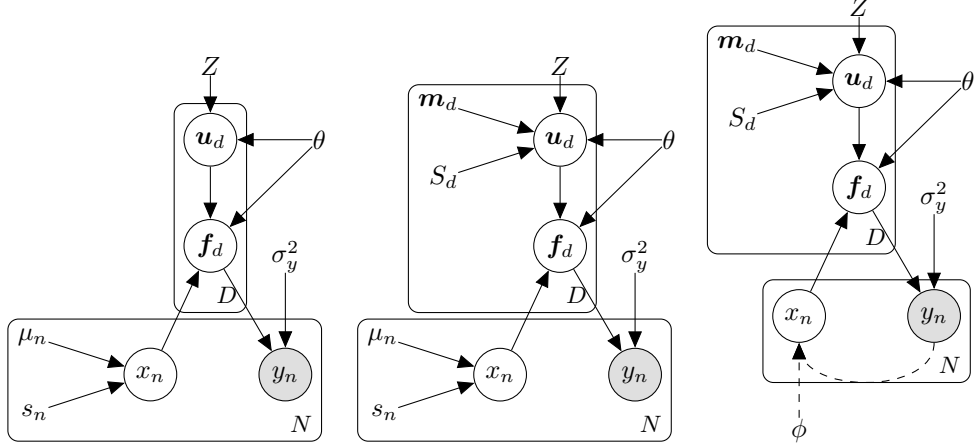
\begin{figure*}
\vspace{-5mm}
\centering
\resizebox{0.75\textwidth}{!}{%
\begin{tikzpicture}[scale=0.08]
\vspace{-3mm}
     \node[const, yshift=0.5cm] (Z) {$Z$};
    \node[latent, below=of Z, yshift=0.5cm] (ud) {$\bm{u}_d$};
    \node[latent, below=of ud, yshift=0.3cm] (fn) {$\bm{f}_{d}$};
    \node[latent, below=of fn, xshift=-0.8cm] (xn) {$x_{n}$};
        \node[obs, below=of fn, xshift=1cm] (yn) {$y_n$};
    \node[const, above=of fn, right=of ud ] (theta) {$\theta$};
    \node[const, above=of yn] (sigma) {{$\sigma^2_{y}$}}; 
    \node[const, left=of xn, yshift=0.5cm] (mun) {$\mu_{n}$};
    \node[const, left=of xn, yshift=-0.5cm] (sn) {$s_{n}$};

   \edge {xn} {fn};
   \edge {fn} {yn};
   \edge {ud} {fn};
   \edge {Z} {ud};
   \edge{mun} {xn};
   \edge {sn} {xn};

   \plate{yf} {(yn)(xn)(mun)} {$N$} ;
    \plate{udp} {(ud)(fn)} {$D$} ;
  \edge {theta} {fn};
   \edge {sigma} {yn};
      \edge {theta} {ud};

\end{tikzpicture}\hspace{3mm}
\begin{tikzpicture}[scale=0.09]
\vspace{-3mm}
    \node[const, yshift=0.5cm] (Z) {$Z$};
    \node[latent, below=of Z, yshift=0.5cm] (ud) {$\bm{u}_d$};
    \node[latent, below=of ud, yshift=0.3cm] (fn) {$\bm{f}_{d}$};
    \node[latent, below=of fn, xshift=-0.8cm] (xn) {$x_{n}$};
        \node[obs, below=of fn, xshift=1cm] (yn) {$y_n$};
    \node[const, above=of fn, right=of ud ] (theta) {$\theta$};
    \node[const, above=of yn] (sigma) {{$\sigma^2_{y}$}}; 
    \node[const, left=of xn, yshift=0.5cm] (mun) {$\mu_{n}$};
    \node[const, left=of xn, yshift=-0.5cm] (sn) {$s_{n}$};
    \node[const, left=of ud, yshift=0.5cm](md){$\bm{m}_{d}$};
    \node[const, left=of ud, yshift=-0.5cm](Sd){$S_{d}$};

   \edge {xn} {fn};
   \edge {fn} {yn};
   \edge {ud} {fn};
   \edge {Z} {ud};
   \edge{mun} {xn};
   \edge {sn} {xn};
   \edge {md}{ud};
   \edge{Sd}{ud};

   \plate{yf} {(yn)(xn)(mun)} {$N$} ;
    \plate{udp} {(ud)(md)(Sd)(fn)} {$D$} ;
  \edge {theta} {fn};
   \edge {sigma} {yn};
      \edge {theta} {ud};
\end{tikzpicture}\hspace{3mm}
\begin{tikzpicture}[scale=0.09]
\vspace{3mm}
     \node[const, yshift=0.5cm] (Z) {$Z$};
    \node[latent, below=of Z, yshift=0.5cm] (ud) {$\bm{u}_d$};
    \node[latent, below=of ud, yshift=0.3cm] (fn) {$\bm{f}_{d}$};
    \node[latent, below=of fn, xshift=-0.8cm] (xn) {$x_{n}$};
    \node[obs, below=of fn, xshift=1cm] (yn) {$y_n$};
    \node[const, above=of yn] (sigma) {{$\sigma^2_{y}$}}; 
    \node[const, above=of fn, right=of ud ] (theta) {$\theta$};
    \node[const, left=of ud, yshift=0.5cm](md){$\bm{m}_{d}$};
    \node[const, left=of ud, yshift=-0.5cm](Sd){$S_{d}$};
    \node[const, below=of xn](phi){$\phi$};

   \edge {xn} {fn};
   \edge {fn} {yn};
   \edge {ud} {fn};
   \edge {Z} {ud};
   \edge {md}{ud};
   \edge{Sd}{ud};
   \path[->,draw,dashed]
  (yn) edge[bend left=90] node [left] {} (xn);
  \draw[->,dashed] (phi) to node {} (xn);

   \plate{yf} {(yn)(xn)} {$N$} ;
    \plate{udp} {(ud)(md)(Sd)(fn)} {$D$} ;
  \edge {theta} {fn};
   \edge {sigma} {yn};
      \edge {theta} {ud};
      
\end{tikzpicture}}
\caption{\emph{Left:} Graphical model of the collapsed bound formulation of the Bayesian GPLVM. \emph{Middle:} B-SVI where we learn individual parameters for each latent point. \emph{Right:} AEB-SVI where the parameters for each latent point are deterministically derived by encoding the data point with the amortising neural network.}
\vspace{-3mm}
\end{figure*}
\vspace{-2mm}
\subsubsection{Analytical derivation of the factorised form: Gaussian and Non-Gaussian likelihoods}
Making the parameterisation of the variational distributions explicit for clarity, we denote the variational distribution over the latent points as $q_{\phi}(\bm{x}_{n})$ where $\phi = \{\mu_{n}, s_{n}\mathbb{I}_{Q}\}$ and the variational distribution over the inducing variables as $q_{\lambda}(\bm{u}_{d})$ where $\lambda = \{\bm{m}_{d}, S_{d} \}$. 
\scalebox{0.90}{
\begin{minipage}{\linewidth}
\begin{align}     \label{elem1}
   \mathcal{L}({\mathcal{D}})
     &= \mathbb{E}_{q(.)}\left[\sum_{n,d}\log \mathcal{N}(y_{n,d};\bm{f}_{d}(\bm{x}_{n}), \sigma^{2}_{y})\right] \\ &- \underbrace{\sum_{n}\textrm{KL}(q_{\phi}(\bm{x}_{n})||p(\bm{x}_{n})) - \sum_{d}\textrm{KL}(q_{\lambda}(\bm{u}_{d})||p(\bm{u}_{d}|Z))}_{\textrm{KL terms}} \nonumber \\ 
    &= \sum_{n, d}\mathbb{E}_{q_{\phi}(\bm{x}_{n})}[\underbrace{\mathbb{E}_{p(\bm{f}_{d}|\bm{u}_{d}, \bm{x}_{n})q_{\lambda}(\bm{u}_{d})}[\log \mathcal{N}(y_{n,d};\bm{f}_{d}(\bm{x}_{n}), \sigma^{2}_{y})]}_{\mathcal{L}_{n,d}(\bm{x}_{n}, y_{n,d}) = \mathcal{L}_{n,d}}]  \nonumber  \\
    & \hspace{50mm} - \textrm{KL terms} \nonumber
\end{align}
\end{minipage}}

The expected log-likelihood term for a single data point $(n)$ and dimension ($d$) -  $\mathcal{L}_{n,d}(\bm{x}_{n}, y_{n,d})$ is reduced to,
\scalebox{0.95}{
\begin{minipage}{\linewidth}
\begin{align}  \label{baseeq}
    &\mathbb{E}_{q_{\phi}(\bm{x}_{n})}[\mathcal{L}_{n,d}] \nonumber \\
   &= \hspace{-2mm} \int q_{\phi}(\bm{x}_{n})\int q_{\lambda}(\bm{u}_{d})\int p(\bm{f}_{d}|\bm{u}_{d}, \bm{x}_{n}) \log\mathcal{N}(y_{n,d};\bm{f}_{d}(\bm{x}_{n}), \sigma^{2}_{y}) \nonumber \\ & \hspace{60mm} d\bm{f}_{d}(\bm{x}_{n})d\bm{u}_{d}d\bm{x}_{n} \nonumber \\
 &= \log \mathcal{N}(y_{n,d}|\underbrace{\langle K_{nm}  \rangle_{q_{\phi}(\bm{x}_{n})}}_{\Psi^{(n,\cdot)}_{1}}K_{mm}^{-1}\bm{m}_{d}, \sigma^{2}_{y}) \nonumber \\
  & -\dfrac{1}{2\sigma^{2}_{y}} \textrm{Tr}(\underbrace{\langle K_{nn} \rangle_{q_{\phi}(\bm{x}_{n})}}_{\psi_{0}^{n}}) + \dfrac{1}{2\sigma^{2}_{y}}\textrm{Tr}(K_{mm}^{-1}\underbrace{\langle K_{mn}K_{nm}
  \rangle_{q_{\phi}(\bm{x}_{n})}}_{\Psi_{2}^{n}}) \\ &
 - \dfrac{1}{2\sigma^{2}_{y}} \textrm{Tr}(S_{d}K_{mm}^{-1}\underbrace{\langle K_{mn}K_{nm}
 \rangle_{q_{\phi}(\bm{x}_{n})}}_{\Psi_{2}^{n}}K_{mm}^{-1}) \nonumber
\end{align}
\end{minipage}}

where we analytically perform the integration w.r.t $q_{\lambda}(\bm{u}_{d})$ and the inner-most integral w.r.t  $p(\bm{f}_{d}|\bm{u}_{d}, \bm{x}_{n})$ leaving behind the expectations w.r.t $q_{\phi}(\bm{x}_{n})$ which are handled numerically with Monte Carlo estimation. 
\scalebox{0.95}{
\begin{minipage}{\linewidth}
\begin{align}
    \Psi^{(n,\cdot)} &\approx \dfrac{1}{J}\sum_{j=1}^{J}k(\bm{x}_{n}^{(j)},Z), \Psi_{2}^{n} \approx \dfrac{1}{J}\sum_{j=1}^{J}  k(Z, \bm{x}_{n}^{(j)})k(\bm{x}_{n}^{(j)}, Z), \nonumber \\ &\hspace{2mm} \psi_{0}^{n} \approx \dfrac{1}{J}\sum_{j=1}^{J} k(\bm{x}_{n}^{(j)}, \bm{x}_{n}^{(j)})
\end{align}
\end{minipage}}
where $\bm{x}_{n}^{(j)} \sim q_{\phi}(\bm{x}_{n})$; the samples $\bm{x}_{j}$ are drawn using the reparameterization trick \cite{kingma2013auto} where we sample $\epsilon^{(j)} \sim \mathcal{N}(0,\mathbb{I}_{Q})$ and $\bm{x}_{n}^{(j)} = \mu_{n} + s_{n}\odot\epsilon^{(j)}$. 
\scalebox{0.9}{
\begin{minipage}{\linewidth}
\begin{align}
     &\mathbb{E}_{q_{\phi}(\bm{x}_{n})}[\mathcal{L}_{n,d}] \simeq  \dfrac{1}{J}\sum_{j=1}^{J}\mathcal{L}_{n,d}(\bm{x}_{n}^{(j)}, y_{n,d}) \nonumber \\ & \simeq  \dfrac{1}{J}\sum_{j=1}^{J}\mathcal{L}_{n,d}(\mu_{n} + s_{n}\odot\epsilon^{(j)}, y_{n,d}) \\
     &= \dfrac{1}{J}\sum_{j=1}^{J}\mathcal{L}_{n,d}(g_{\phi}(\epsilon^{(j)}), y_{n,d}) \nonumber
\end{align}
\end{minipage}}
We denote the approximate ELBO as $\mathcal{\hat{L}}$,
\begin{align}
\mathcal{\hat{L}} &= \sum_{n}\sum_{d} \overbrace{\dfrac{1}{J}\sum_{j=1}^{J}\mathcal{L}_{n,d}(g_{\phi}(\epsilon^{(j)}), y_{n,d})}^{\mathcal{\hat{L}}_{n,d}} \\ & \hspace{-7mm} - \sum_{d}\textrm{KL}(q_{\lambda}(\bm{u}_{d})||p(\bm{u}_{d}|Z)) - \sum_{n}\textrm{KL}(q_{\phi}(\bm{x}_{n})||p(\bm{x}_{n})) \nonumber
\end{align}
For a non-Gaussian likelihood (following on from \cref{baseeq}), the expectations around the log-likehood term are intractable, instead one simplifies down to the marginals $q(\bm{f}_{d}|\bm{x}_{n})$ analytically computable with standard Gaussian identities,
\begin{align} \label{nong}
    &\int p(\bm{f}_{d}|\bm{u}_{d}, \bm{x}_{n})q_{\lambda}(\bm{u}_{d})d\bm{u}_{d} = q(\bm{f}_{d}|\bm{x}_{n})  \\
    &= \mathcal{N}(k_{n}^{T}K_{mm}^{-1}\bm{m}_{d}, k_{nn} + k_{n}^{T}K_{mm}^{-1}(S_{d} - K_{mm})K_{mm}^{-1}k_{n}) \nonumber
\end{align}
where $k_{n}^{T}$ is the $n^{th}$ row of $K_{nm}$ only dependent on input $\bm{x}_{n}$ and $k_{nn} = k(\bm{x}_{n}, \bm{x}_{n})$. Further, $q(\bm{f}_{d}|\bm{x}_{n})$ denotes the marginal latent GP $\bm{f}_{d}$ conditioned at input $\bm{x}_{n}$. This gives the simplified lower bound,
\begin{align} \label{final_elbo}
    \mathcal{\hat{L}} &= \sum_{n,d}\mathbb{E}_{q(\bm{f}_{d}|\bm{x}_{n})q_{\phi}(\bm{x}_{n})}[\log p(y_{n,d}|\bm{f}_{d}(\bm{x}_{n}))] \\ & - \sum_{d}\textrm{KL}(q_{\lambda}(\bm{u}_{d})||p(\bm{u}_{d}|Z)) - \sum_{n}\textrm{KL}(q_{\phi}(\bm{x}_{n})||p(\bm{x}_{n})) \nonumber 
\end{align}
The \textsc{point} model in experiments comprises of just the first two terms in \cref{final_elbo}, while the MAP method excludes the KL divergence term for latents ($\bm{x}_{n}$) in exchange for solely the prior term $p(\bm{x}_{n})$. Finally, in order to speed-up computation we use mini-batches (see Algorithm 1) to construct a scalable, differentiable and unbiased estimator optimised with standard stochastic gradient methods. The KL terms are analytically tractable due to the choice of the Gaussian variational family for $q_{\phi}(\bm{x}_{n})$ and the optimal (Gaussian) variational family for $q_{\lambda}(\bm{u}_{d})$.

The method is known as \textit{doubly stochastic variational inference} due to the two-fold stochasticity attributed to computing numerical expectations by sampling from the variational distributions $q(\bm{x}_{n})$ and due to mini-batching for gradient updates.

\begin{algorithm*}
\SetAlgoLined
\label{algorithm}
\footnotesize
\caption{Bayesian GPLVM with Doubly Stochastic Variational Inference (\textbf{B-SVI})}
\vspace{2mm}
\textbf{Input:} ELBO objective $\mathcal{L}$, gradient based optimiser \texttt{optim()}, training data $\mathcal{D} = \{\bm{y}_{n}\}_{i=1}^{N}$ \\ Initial model params: \\
\quad $\bm{\theta}$ (covariance hyperparameters for GP mappings $\bm{f}_{d}$ and data noise variance $\sigma^{2}_{y}$),\\
 \vspace{1mm}
Initial variational params: \\
\quad $Z \in \mathbb{R}^{M \times Q}$ (inducing locations), \\
\quad $\phi = \{\mu_{n}, s_{n} \}_{n=1}^{N}$ (local variational parameters -  $\bm{x}_{n} \sim \mathcal{N}(\mu_{n},s_{n}\mathbb{I}_{Q}), \mu_{n}, s_{n} \in \mathbb{R}^{Q}$ ) \\
\quad $\lambda = \{m_{d}, S_{d} \}_{d=1}^{D}$ (global variational parameters - $ \bm{u}_{d} \sim \mathcal{N}(m_{d}, S_{d}), \bm{u}_{d} \in \mathbb{R}^{M}, S_{d} \in \mathbb{R}^{M \times M}$ ) \\
\vspace{2mm}
\While{not converged}{ 
\begin{itemize}
  \setlength{\itemsep}{0pt}
   \item Choose a random mini-batch $\mathcal{D}_{B} \subset \mathcal{D}$. \\
   \item Sample $J$ samples from the noise distribution $\epsilon^{(j)} \sim \mathcal{N}(0,\mathbb{I}_{Q})$. \\
   \item Form a mini-batch estimate of the ELBO: \\
   \qquad $\mathcal{\hat{L}}(\mathcal{D}_{B}) = \dfrac{N}{B}\left(\sum_{b}\sum_{d}\mathcal{\hat{L}}_{b,d} - \sum_{b}\textrm{KL}(q_{\phi}(\bm{x}_{b})||p(\bm{x}_{b})\right) - \sum_{d}\textrm{KL}(q_{\lambda}(\bm{u}_{d})||p(\bm{u}_{d}|Z)) $ \\
   \item Gradient step: $Z, \bm{\theta}, \sigma^{2}_{y}, \{\mu_{b}, s_{b} \}_{b=1}^{B} \{m_{d}, S_{d} \}_{d=1}^{D} \longleftarrow $ \texttt{optim}$(\mathcal{\hat{L}}(\mathcal{D}_{B}))$
\end{itemize}
}
\Return{$Z, \bm{\theta}, \phi, \lambda $}
\end{algorithm*}

\vspace{-2mm}

\subsection{Amortised Inference with Encoders}

The GPLVM model provides a probabilistic non-linear mapping from latent space $\bX$ to data space $\bY$, hence,  local distances are preserved in the latent space ensuring that points \textit{close}\footnote{For a stationary kernel, this would be closeness in a sense of Euclidean distance.} in latent space recover observations that are close in data space. \cite{lawrence2006local} and \cite{bui2015stochastic} additionally account for this feature of data distance preservation by introducing an encoder within the GPLVM model (see also \citep{dai2015variational}). 


\textsc{AEB-SVI:} In this variational model, the mean and variance of the base Gaussian distribution are parameterised as outputs of individual neural networks $G_{\phi_{1}}$ and $H_{\phi_{2}}$ with network weights $\phi_{1}$ and $\phi_{2}$. The network weights are shared across all the data points enabling amortised learning \citep{bui2015stochastic}. The key property of this parameterisation is that it learns a dense covariance matrix (parameterised through a factorization) per data-point thereby capturing correlations across dimensions (per latent point) in latent space. 
\scalebox{0.95}{
\begin{minipage}{\linewidth}
\begin{equation}
q(\bX) = \prod_{n=1}^{N}\mathcal{N}( \bm{x}_{n}; G_{\phi_{1}}(\bm{y}_{n}), H_{\phi_{2}}(\bm{y_{n}}) H_{\phi_{2}}(\bm{y_{n}})^{T}) 
\label{nn_bc}
\end{equation}
\end{minipage}}
This function is usually referred to as the back-constraint and its parameters are \textit{global}, i.e. shared between all the data points. This allows for fast amortised inference and constant time test predictions. \cite{bui2015stochastic} present this model for a Gaussian likelihood setting.

\subsection{Predictions}

When unseen high-dimensional points arrive in data space $\bm{y}^{*}$ we are interested in computing the latent point distribution $q(\bm{x}^{*})$ per test point $\bm{y}^{*}$ where we have access to the trained variational parameters $(\phi, Z, \lambda)$ and model hyperparameters ($\bm{\theta}$). One motivation for auto-encoder driven models is that we have constant-time $\mathcal{O}(1)$ test predictions. Given a test point $\bm{y}^{*}$, we use the set of global encoder weights ($\phi_{1}, \phi_{2}$) to obtain the posterior approximation $q(\bm{x}^{*})$ (as in eq. \ref{nn_bc}). In the \textsc{B-SVI} model (Algorithm 1.) we can't obtain the distributional parameters for $q(\bm{x}^{*})$ deterministically, instead we re-optimise the ELBO with the additional test data point $\bm{y}^{*}$ while keeping all the global and model hyperparameters frozen at their trained values. Note that since the SVI ELBO factorises across data points, $\mathcal{L}(\{\bm{y}_{n}\}_{n=1}^{N}, \bm{y}^{*}) = \sum\limits_{n=1}^{N+1}\sum\limits_{d=1}^{D} \mathcal{L}_{n,d}$, the gradients to derive the distributional parameters of the test point $\mathcal{N}(\mu_{*}, s_{*}\mathbb{I}_{Q})$ only depend on the component terms.  

\begin{figure*}[t]
    \includegraphics[scale=0.57]{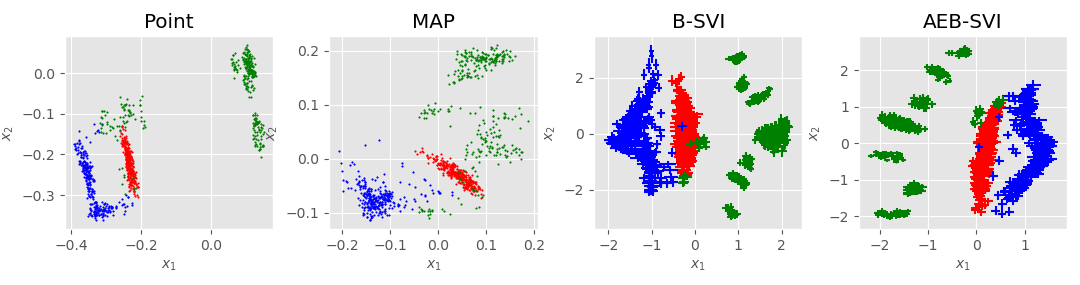} \\
    \includegraphics[scale=0.57]{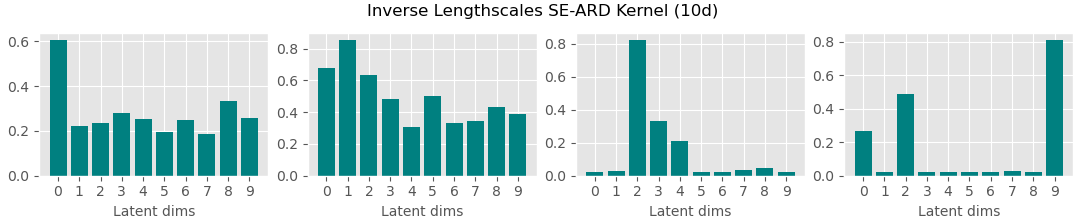}
    \caption{\small{Top: The 2$d$ latent space corresponding to the dominant dimensions learnt by each model. Bottom: The inverse lengthscales learnt by each model specification. We include a similar report for qPCR in the supplementary.}}
    \label{fig:oilflow}
\end{figure*}

\subsection{Computational Complexity}

The training cost of the canonical non-SVI Bayesian GPLVM  is dominated by $\mathcal{O}(NM^{2}D)$ where $M << N$ is the number of inducing variables and $D$ is the data-dimensionality (we have $D$ GP mappings $\textbf{f}_{d}$ per output dimension), with the SVI framework this is reduced to $\mathcal{O}(M^{3}D)$ (free of $N$). The practical algorithm is made further scalable with the use of mini-batched learning admissable under the uncollapsed lower bound (this work). However the number of global variational parameters to be updated in each step (parameters of $q(\textbf{U})$) is now increased. We summarise the number of global and local variational parameters across all the latent variable formulations in the table \ref{compute}.
\begin{table}[h]
\vspace{2mm}
    \caption{\small{Summary of compute across different models.}}
    \scalebox{0.75}{
    \begin{minipage}{\linewidth}
    \begin{tabular}{c|c|c|c|c|c}
         Model & Canonical & \textsc{Point} & \textsc{MAP} & \textsc{B-SVI}   & \textsc{AEB-SVI} \\
         \hline 
        Global ($\lambda$) & $MQ$ & \multicolumn{3}{|c|}{$MQ + MD + M^{2}D$} & $MQ$ + $|\phi_{1}+\phi_{2}|$\\
        \hline
        Local & $2NQ$ &  \multicolumn{2}{|c|}{$NQ$} & $2NQ$ & -- 
    \end{tabular}
      \end{minipage}}
    \label{compute}
    \vspace{-5mm}
\end{table}
The `Canonical' model refers to \citet{titsias2010bayesian} and depends on the optimisation of $MQ$ global parameters pertaining to the $Q$-dimensional inducing inputs $Z$. \textsc{B-SVI} on the other hand depends on $MQ + MD + M^{2}D$ parameters (inducing inputs, mean and dense covariance of inducing variables per latent dimension). The \textsc{AEB-SVI} model only has global parameters. The number of local variational parameters (parameters of $q(\textbf{X})$) are the same between the canonical and B-SVI model at $2NQ$. At prediction time we need to learn the $2N^{*}Q$ local variational parameters from the augmented ELBO (for both the canonical and B-SVI model), this is further sped up in our framework with the AEB-SVI model which provides constant time $\mathcal{O}(1)$ test predictions.

\subsection{Training with Many Missing Dimensions}

A key motivation for our framework is dealing with missing data at \emph{training time}. Most machine learning algorithms are designed to be deployed on carefully curated tables of data with a fixed number of features. If data is missing, it is often dealt with through EM algorithms which can deal with missingness up to around 30\%. In the real world the situation is often very different. Important data sets such as electronic health records can have 90\% or more missing values. In these domains the objective function becomes dominated by the missing values and learning fails to occur \citep{Corduneanu-continuation02}. We consider a data set-up where every vector $\bm{y}$ has an arbitrary number of dimensions missing and there is no constraint or structure about their \textit{missingness}. Our training procedure leverages the marginalisation principle of Gaussian distributions and the fact that the data dependent terms of the SVI ELBO factorise across data points and dimensions. This means we can trivially marginalise out the missing dimensions $\bm{y}_{a}$, because each individual data point $\bm{y}$ is modelled as a joint Gaussian. Consider a high-dimensional point $\bm{y}$ which we split into observed, $\bm{y}_o$ and unobserved $\bm{y}_a$ dimensions, 
\scalebox{0.89}{
\begin{minipage}{\linewidth}
\begin{equation}
    \int \prod_{d \in a} \prod_{d \in o} p(\bm{y}_{a}, \bm{y}_{o} | \bm{u}_{d}, \bX) d\bm{y}_{a} = \prod_{d \in o} p(\bm{y}_{o} | \bm{u}_{d}, \bX)
\end{equation}
  \end{minipage}}
where $a$ and $o$ denote the indices of missing and observed dimensions respectively and all dimensions are given as, $D = {a \cup o}$. $\bm{u}_{d} \in \mathbb{R}^{M}$ denote the inducing variables which ensure conditional independence. The latent variables per data point $\bm{x}_{n}$ are informed by the observed dimensions only, while the $M$ inducing variables per dimension $\bm{u}_{d}s$ are informed by all the data points which have the observed dimension. The elegance of this framework is that there is no major change in the training procedure as the ELBO eq. \ref{final_elbo} sums over all observed dimensions per data point. We can also easily reconstruct the missing training dimensions by decoding the mean of the optimised variational latent distribution $q(\bm{x}) = \mathcal{N}(\mu^{*} s^{*}\mathbb{I}_{Q})$.

This set-up reflects real-world data which is often sparse with many missing and few overlapping dimensions across the full dataset. The experiments in section \ref{partial} demonstrate the reconstruction ability of \textsc{B-SVI} when faced with missing dimensions at training time. The missing data framework is not immediately compatible with auto-encoding models (\textsc{AEB-SVI}) as every latent point $\bm{x}_{n}$ is expressed as a function of the data point $\bm{y}_{n}$. However, set encoders \citep{qi2017pointnet,vedantam2017generative,ma2018eddi} can be integrated as the auto-encoding component instead of a standard neural network. We defer this to future work.  
\section{Experiments}

\begin{table*}[t]
\caption{\small{Test RMSE for datasets with $\pm$ standard error across 3 optimisation runs. $Z$ denotes the number of inducing variables used per dimension and $Q$ denotes the dimension of the latent space.}}
\label{summary}
\scalebox{0.92}{
\begin{tabular}{c|c|c|c|c|c|c|c|c}
Dataset & $N$ / $d$ & Likelihood & $Z$ & $Q$ & \textsc{Point} & \textsc{MAP}  & \textsc{B-SVI} & \textsc{AEB-SVI}  \\
\hline
Oilflow & 1000 / 12 &  Gaussian & 25 & 10 & 0.341 (0.008) &  0.569 (0.092) & 0.0925 (0.025) & \textbf{0.067 (0.0016)} \\
qPCR & 450 / 48 &   Gaussian & 40 & 11 & 0.624 (0.027) & 0.589 (0.016)  & \textbf{0.554 (0.017)} & \textbf{0.539 (0.004)} \\
Taxi-cab & 744 / 3 & Poisson & 36 & 2 & 118 (21) & 134 (11) & 249 (81) & 232 (22)
\end{tabular}}
\vspace{-3mm}
\end{table*}
\subsection{Ablation Study: Quantitative Results}
\label{ablation}
\textbf{Models:} Our experiments implement four incarnations of the GPLVM model namely, \textsc{Point} which refers to the Sparse GPLVM, \textsc{MAP} which refers to the sparse GPLVM with a prior over latent variables $\bm{x}_{n}$, the Bayesian SVI model \textsc{B-SVI} and \textsc{AEB-SVI} which refers to the Autoencoded Bayesian GPLVM. We assess each model on their ability to reconstruct unseen high-dimensional points, automatic regularisation and detecting class structure in latent space. Further results and full details about the experimental set-up are enclosed in the supplementary material.

\textbf{Data set-up:} The multi-phase Oilflow data \citep{bishop1993analysis} consists of 1000, 12$d$ data points belonging to three classes which correspond to the different phases of oil flow in a pipeline. The qPCR data contains 48 dimensional single-cell data obtained from mice \citep{guo2010resolution} where each dimension corresponds to a gene. Cells differentiate during their development and these data were obtained at various stages of development which contribute 10 categories/classes to which each of the cell belongs. We also use a count dataset constructed from the NYC taxi cab records \citet{taxi} where we use vehicle counts of yellow/green/for-hire cabs aggregated by hour over the month of Jan 2020. We use a 80/20 split for training/testing and report test performance with $\pm$ 2 standard errors over three optimization runs. Since the training is unsupervised, the inherent ground-truth labels were not a part of training. 

The 2$d$ projections of the latent space (for oilflow data) clearly show that all variants are able to discover the class structure. It is important to note that unlike previous versions these models do not require PCA initialisation and all models were initialised randomly. In order to highlight certain features, the latent dimensionality ($Q$) was kept fixed across all models.

\textsc{Point} and \textsc{MAP} overfit as can be seen from the magnitude of the inverse lengthscales across all the latent dimensions. Both  \textsc{Point} and \textsc{MAP} find all the latent dimensions relevant. Conversely, \textsc{B-SVI} and \textsc{AEB-SVI} identify two or three dominant dimensions to represent the data exhibiting automatic regularisation along with better test reconstruction errors.

\begin{figure}[h]
    \includegraphics[scale=0.45]{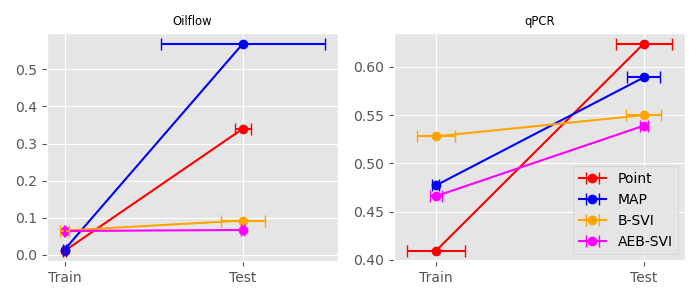}
    \caption{Left: The train and test RMSE per model showing evidence on overfitting for the non-Bayesian incarnations of Point and MAP.} 
    \label{fig:analysis}
\end{figure}

The training/test error comparison (fig. \ref{fig:analysis}) provides further evidence of overfitting in the point methods for high-dimensional datasets. The quality of the 2$d$ latent projection of training data using the fully trained model might hide the overfitting effects as it is equally effective at disentangling the class structure. 

For the taxi-cab data we use the likelihood, $p(\bm{y}_{n}|\bm{f}_{n}) = \prod_{d=1}^{D}\textrm{Poisson}(\exp(\bm{f}_{d}(\bm{x}_{n})))$ and 10 samples from $q(\bm{x}_{n})$ to approximate the expectation. All methods give very good test reconstructions (see supplementary for plots), however, it might seem like \textsc{B-SVI} and \textsc{AEB-SVI} underperform due to the higher RMSEs but the magnitude of the count values lead to larger variations in the test scores reported. An important factor is the dimensionality of the data, the benefit of the Bayesian techniques are subdued when acting on low-dimensional data as well as the importance of capturing correlations in latent space is more pronounced when the data has several dimensions. We show additional analysis and reconstructions in the supplementary where the Bayesian methods with SVI don't overfit even when we match the latent space dimensionality to that of the data space.


\subsection{Missing data: Reconstructing Structured Images \& Human Motion}
\label{partial}
\begin{figure*}[h]
\centering
\includegraphics[scale=0.37]{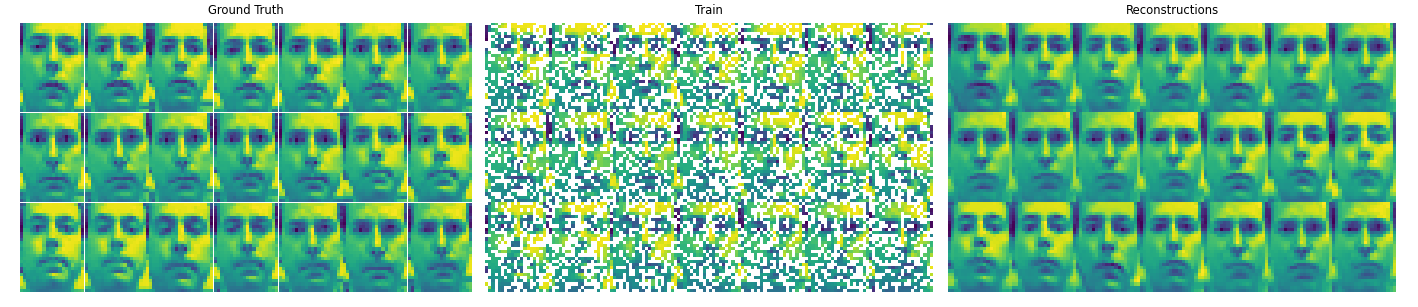}
\includegraphics[scale=0.66]{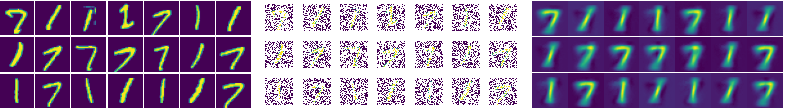}
\caption{\small{Top Row: Brendan faces reconstruction task with 39$\%$ missing pixels. Bottom row: MNIST reconstruction task where the digits were trained on partially observed images. In both rows the left column denotes ground truth data, the center column denotes a subset of the training data and the right column denotes reconstructions from the 5d latent distribution for MNIST and Brendan respectively.}}
\label{recon}
\end{figure*}

The focus of this experiment is to qualitatively assess how the models capture uncertainty when training with missing data in structured inputs. We use 15K training samples from the MNIST digits dataset \citep{lecun2010mnist} with $\approx$ 60$\%$ of the pixels missing at random in each digit. Each image has 768 pixels yielding a 768$d$ data space.  The image data set \citep{roweis2000nonlinear} contains $\approx$2000 images of a face taken from sequential frames of a short video. Each image is of size 20x28 yielding a 560$d$ data space. Fig. \ref{recon} summarises sample generation from the learnt 5$d$ latent distribution. Note that this reconstruction experiment differs from the less challenging \emph{test-time} missing data which has been demonstrated in related work \citet{titsias2010bayesian,gal2014distributed}.
\begin{figure}
    \centering
    \includegraphics[scale=0.55]{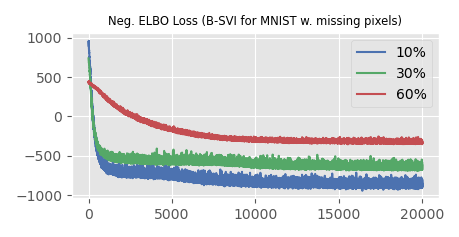}
    \includegraphics[scale=0.53]{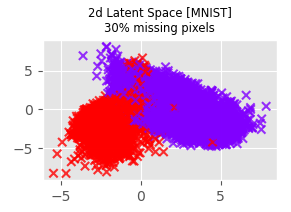}
    \includegraphics[scale=0.53]{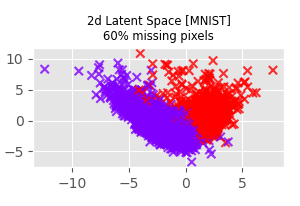}
    \caption{\small{Top: ELBO loss for training with different degrees of missing pixels. Bottom: 2d latent space corresponding to the smallest lengthscales, note it is possible to disentangle the 2 digit classes quite effectively with almost no degradation for double the fraction of missing pixels.}}
    \label{fig:elbos}
\end{figure}




\begin{table*}
\centering
\caption{\small{Test NLPD for datasets with $\pm$ standard error across 5 optimisation runs. The NLPDs across formulations indicate superior uncertainty quantification for the Bayesian schemes for both full and missing data problems.}}
\label{summary_nlpd}
\scalebox{0.87}{
\begin{minipage}{\linewidth}
\begin{tabular}{c|c|c|c|c|c}
Dataset & \multicolumn{4}{|c|}{Test NLPD} &  Missing($\%$) \\
\hline 
 & \textsc{Point} &  \textsc{MAP}  &  \textsc{B-SVI (Ours)} & \textsc{AEB-SVI} & \\
\hline
Oilflow &  4.104 (3.223) &  8.16 (1.224) & \textbf{-11.3105 (0.243)} & \textbf{-11.392 (0.147)} & -- \\
qPCR &   32.916 (3.39) & 30.899 (2.399) & \textbf{27.844 (1.429)} & \textbf{25.422 (2.004)} & --  \\
MOCAP & 35472.566 (445.82) & 8904.162 (162.45) & \textbf{2275.021 (33.89)} & --  & 44.5$\%$  \\
\end{tabular}
\end{minipage}}
\end{table*}

\begin{table}[h]
\caption{\small{Test RMSE for training with different degrees of missing dimensions per datapoint. The quality of reconstruction is best when the $\%$ missing during training matches the fraction of missing dimensions during testing.}}
\label{missing_summary}
\scalebox{0.85}{
\begin{tabular}{c|c|c|c|c|c}
Dataset & \multicolumn{2}{c|}{$\%$ missing ($\rightarrow$ train $\%$)} & 10$\%$  &  30$\%$  & 60$\%$ \\
\hline
 \multirow{3}{*}{MNIST} & \multirow{3}{*}{($\downarrow$ test $\%$)} & 10$\%$ &  0.2716 & 0.2735 & 0.2763   \\
 &  & 30$\%$ &  0.2731 & 0.2730 & 0.2794  \\
 &  & 60$\%$ & 0.2755 & 0.2762 & 0.2748 \\
 \hline 
\end{tabular}}
\end{table}

\begin{figure}[h]
\centering
\includegraphics[scale=0.50]{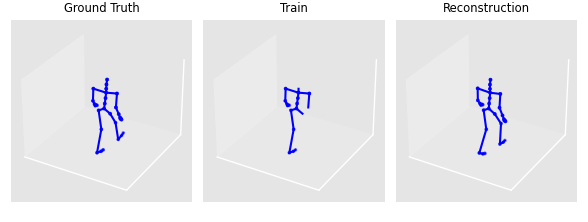}
\includegraphics[scale=0.43]{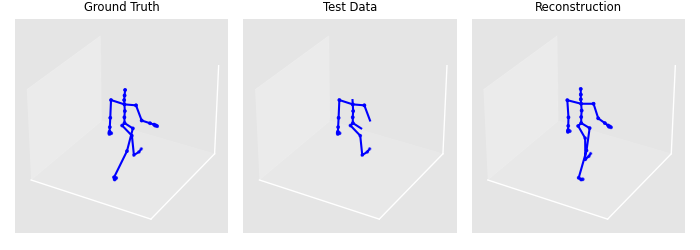}
\caption{\small{The train and test reconstruction of a single high-dimensional human pose (top: walking) and (bottom: running). The whole training exercise was conducted on incomplete silhouettes to extrapolate to sensible human poses at test time. For instance, the hand location during running was missing in this particular instance and was reconstructed to a remarkable similarity with the ground truth. We include several reconstructions in the supplementary.}}
\label{fig:mocap}
\end{figure}

\begin{table}
\vspace{2mm}
\caption{\small{The run time-comparisons highlight the important caveat that the amortised scheme (despite $\mathcal{O}(1)$ test-time predictions) is much slower (2x) as the encoder weights are global variational parameters which are updated at every mini-batch iteration as opposed to B-SVI.}}
\label{summary_runtime}
\scalebox{0.87}{
\begin{minipage}{\linewidth}
\centering
\begin{tabular}{c|c|c|c|c}
Dataset & \multicolumn{4}{|c}{Avg. Iterations/sec.}\\
\hline
 & \textsc{Point} &  \textsc{MAP}  &  \textsc{B-SVI (Ours)} & \textsc{AEB-SVI} \\
\hline
Oilflow &   167.56 & 168.42 & 164.43 & 89.42 \\
qPCR &   133.59 & 126.41 & 113.85 & 54.61 \\
MOCAP &  161.72 &  159.64 & 140.79 & -\\
\end{tabular}
\end{minipage}}
\vspace{-2mm}
\end{table}

To demonstrate the versatility of the reconstruction task we tested the method on several examples of the \textit{walking}, \textit{jumping} and \textit{running} human pose from the CMU motion capture database. We split up these motions into four sections, and remove an assortment of body components. We then try to recreate the entire body movement using the \textsc{B-SVI} formulation. A sample reconstruction for a training point and a test point is shown in fig. \ref{fig:mocap}.

\subsection{MovieLens100K}
\begin{figure}[h]
    \includegraphics[scale=0.57]{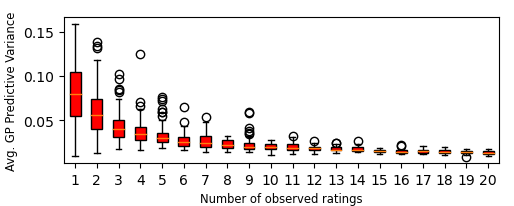}
    \caption{\small{GP predictive variance around the ratings for a movie as a function of how many users have rated the movie. The higher the number of times a movie has been rated, the less uncertain is its prediction and vice-versa.}}
    \label{fig:ml}
\end{figure}

The movie lens 100K data has 1682 movies (columns/$D$) across 943 users (rows/$N$) where each user has rated an average of 20 movies \citep{harper2015movielens}. The ratings range from $\{1,2,\ldots,5\}$. This yields an extremely sparse data grid with $93.8\%$ of the entries missing.\footnote{each row denotes a user, when a user has not rated a movie the value is NaN.} We learn a $10d$ latent distribution for the movie lens data and assess the quality of uncertainty estimates (for the reconstructed ratings) obtained with the \textsc{B-SVI} model (see fig. \ref{fig:ml}).
\section{Related Work}
\textbf{GPLVM \& Variants:} The GPLVM model has spawned several variants since its introduction in \cite{lawrence2004gaussian}. The most fundamental variants are summarised in table \ref{pastwork}. Apart from these there has been a suite of work extending the canonical Bayesian GPLVM model to target different objectives. \citep{damianou2016variational} provides a rigorous examination of the evidence lower bound in the Bayesian GPLVM formulation and extends it to multiple scenarios which include high-dimensional time-series \citep{damianou2011variational} and uncertain inputs for GP regression. The shared GPLVM model \citep{ek2007gaussian} considers a generative model with multiple sources of data and learns a shared representation in the latent space, capable of generating data in the joint observation space. \citep{gal2014distributed} reformulate the Bayesian GPLVM enabling a distributed inference algorithm. \citet{urtasun2007discriminative} use GPLVMs in the context of classification using discriminative priors in latent space and \cite{urtasun2008topologically} focus on embedding data in non-Euclidean latent spaces which is useful when high-dimensional data lie on a natural manifold, e.g. human motion. Other relevant works include \citep{dai2015variational} which augment a deep GP with a recognition model for latent variable inference. None of these works use SVI for inference in these models.

\textbf{VAEs:} Deep probabilistic generative models like VAEs \citep{kingma2013auto} represent a related class of models where the decoder is a parameterised neural network. They have been hugely popular as an unsupervised learning tool for modelling images, large-scale object segmentation and frequently rely on convolutional neural nets as part of the encoding architecture. The most prominent variants include \citep{higgins2016beta}, \citet{kim2018disentangling}, and \citep{sohn2015learning} which focus on disentanglement in latent space as a way to target superior output reconstruction. Structured VAEs need a large amount of input data to train and are unsuitable for tasks with only a moderate sized datasets (such as those used in the ablation study). 
\vspace{-2mm}
\section{Conclusion}
This paper introduces a generalised inference strategy for GPLVM models with key properties of scalable inference, flexible latent variable formulations, likelihoods and most importantly the ability to handle missing data during training. The non-parametric nature of the Gaussian process decoder makes this framework unique to deep parameteric latent variable models like VAEs \citep{kingma2013auto} and allows for robust, interpretable uncertainty around the predictions. A key characteristic of our model is its ability  to train in the \emph{massively missing data} regime that is inadequately addressed by modern parametric machine learning models. We showed in experiments that a fully Bayesian training procedure in conjunction with SVI yields excellent test time performance in these settings. The approach  can be seamlessly extended to learn richer variational families in latent space along with missing data. Future work would focus in that direction.

\subsection*{Acknowledgements}
We thank Carl Henrik Ek and Eric Bodin for feedback on earlier versions of this manuscript. We also thank anonymous reviewers for their comments and feedback which helped clarify the message of the manuscript. VL acknowledges funding from the Qualcomm Innovation Fellowship (EU).
\bibliography{ref}


\clearpage
\appendix

\thispagestyle{empty}

\onecolumn \makesupplementtitle
\section{Broader Impact $\&$ Limitations}
This work contributes a scalable method of inference for Bayesian GPLVM models used for non-parametric, probabilistic dimensionality reduction. Unsupervised learning tasks involving high-dimensional data are ubiquitous in the modern world. Some concrete examples are single-cell RNA data, financial time-series and medical records. In terms of applications, the GPLVM has been widely used in the biological sciences \citep{ahmed2019grandprix}, \citep{verma2020robust} and engineering domains, with the most prominent applications in microarray qPCR datasets to infer the evolution of branching structure in genes \citep{campbell2015bayesian}. One can identify structure in the high-dimensional data by analysing the clustering of low-dimensional latent factors. In the last few years there has been a proliferation of probabilistic generative models using deep neural networks like variational auto-encoders and variants which work extremely well on large and structured datasets, however canonical Bayesian GPLVM models \citep{titsias2010bayesian} originally worked best on small to moderate sized datasets. With the introduction of \textsc{B-SVI} in this work we further extend their domain to larger datasets. Further, the reason they adapt well to smaller datasets comes down to the non-parametric nature of Gaussian processes. Since these models concern non-parametric and probabilistic dimensionality reduction we believe these models can be useful in a much broader range of problems. Further, the fact that these models can train in the presence of missing data is a significant advantage and several real world datasets like medical records, corrupted images and ratings data are only partially observed. There is no straightforward way to deal with missing data in parametric models. Some important pitfalls to keep in mind when training with these models is the difficulty of assessing convergence and the variance of the doubly stochastic ELBO. It is important to ensure that the parameters of the latent distributions have converged, further one must carefully tune experimental parameters like the combination of batch-size and learning rate to achieve optimal performance. 

\section{Relationship to \citep{murray2018mixed}}

\citet{murray2018mixed} use the non-back constrained model with SVI along with non-Gaussian likelihoods but the scope of their experiments is limited to small datasets (max dimension 80) and assess the quality of clustering in the latent space comparing across likelihoods. They do experiment with missing values by dropping some attributes from ~20$\%$ of the data, this means that the model can still sees full ground-truth on the remaining $80\%$ of the data. This is very different to our framework where we conduct a systematic study of robustness of the model when training in the presence of massively missing data. We study the case when all training data is incomplete and has a high $\%$ of randomly missing attributes in each point, training and test (we show high fidelity reconstructions for MNIST and MOCAP which can be seen in further results section of the supplementary). An important point is that the testing/predictive framework summarised in section 3.4 (main paper) has not been effectively explored in more recent literature as the non-trivial setting requires a re-optimisation of the augmented ELBO to learn the latent points of the unseen $\textbf{y}^{*}.$ This is one of the reasons a lot of literature, for instance, \citet{ramchandran2021latent, bui2015stochastic} resort to the amortised set-up.

\section{Theory $\&$ Derivations}
\label{derivations}

\subsection{Motivation for inducing variables}

The sparse inducing variable formulation is integral to the tractability of the Bayesian GPLVM. In order to see this, 
we proceed to derive a lower bound without inducing variables. As is standard, we wish to minimize the KL divergence between the variational approximation and the true posterior given by, $ \textrm{KL}(q(\{\bm{f}_{d} \}_{d=1}^{D}, \bX)||p(\{\bm{f} \}_{d=1}^{D}, \bX|\bY))$. Collecting all the $\bm{f}_{d}$'s in $\bF$ for ease of notation.

\begin{align}
    \kl(q(\bF, \bX)||p(\bF, \bX|\bY)) &= \int q(\bF, \bX) \log \dfrac{q(\bF, \bX)}{p(\bF, \bX|\bY)}d\bF d\bX \\
    &= - \underbrace{\int q(\bF, \bX) \log \dfrac{p(\bY| \bF, \bX)\red{p(\bF|\bX)}p(\bX)}{q(\bF, \bX)} d\bF d\bX}_{\textrm{ELBO}} + \log p(\bY)
\end{align}
The evidence lower bound shown above is mathematically and computationally intractable due to the term $p(\bF | \bX) = \prod_{d=1}^{D} p(\bm{f}_{d}|\bX) = \prod_{d=1}^{D}\mathcal{N}(\bm{0}, K_{nn}^{(d)})$ involving the variables $\bX$ which appear non-linearly in the kernel matrix. The augmented bound constructed with inducing variables $\bm{u}_{d}$ for each dimension circumvents this intractability by leading to the cancellation of the difficult term in red.\\

\subsection{Derivation of the DS-ELBO}

We introduce auxiliary inducing variables, $\bm{u}_{d} \in \mathbb{R}^{M}$ for each of the latent functions $\bm{f}_{d}$. Variational inference in the augmented $(\bF, \bU, \bX)$ space is tractable. 

The augmented variational approximation,

\begin{equation}
p(\bF,\bU, \bX|\bY) \approx q(\bF, \bU, \bX) = \prod_{d=1}^{D}[p(\bm{f}_{d}|\bm{u}_{d}, \bX)q(\bm{u}_{d})]\prod_{n=1}^{N}q(\bm{x}_{n}) 
\end{equation}

leads to the following KL between the approximation and the true posterior,

\begin{align*}
     \kl(q(\bF,\bU, \bX)||p(\bF,\bU, \bX|\bY)) &= \int p(\bF|\bU, \bX)q(\bU)q(\bX) \log \dfrac{ p(\bF|\bU, \bX)q(\bU)q(\bX)}{p(\bF, \bU, \bX|\bY)}d\bF d\bU d\bX \\
     &= - \int p(\bF|\bU, \bX)q(\bU)q(\bX) \log \dfrac{p(\bY|\bF, \bX){\cancel{\color{red}{p(\bF|\bU, \bX)}}}p(\bU|Z)p(\bX)}{{\cancel{\color{red}{p(\bF|\bU, \bX)}}}q(\bU)q(\bX)}d\bF d\bU d\bX \\
      & \hspace{90mm} + \log p(\bY)
\end{align*}
The final ELBO is given by,

\begin{equation}
\mathcal{L} = \int p(\bF|\bU, \bX)q(\bU)q(\bX) \log \dfrac{p(\bY|\bF, \bX)p(\bU|Z)p(\bX)}{q(\bU)q(\bX)}d\bF d\bU d\bX
\label{elbo2}
\end{equation}

\subsection{Derivation of the expected likelihood term eq. (6)}

In this section we explicitly tackle the triple integration in the expected likelihood term.

\begin{align}
\hspace{-3mm}
    \mathcal{L}_{1} &= \sum_{n,d} \mathbb{E}_{p(\bm{f}_{d}|\bm{u}_{d}, \bX)q(\bm{u}_{d})q(\bm{x}_{n})}[\log p(y_{n,d}|\bm{f}_{d}, \bm{x}_{n})] \\
    &= \sum_{n,d} \int q(\bm{x}_{n}) \int q(\bm{u}_{d}) \underbrace{\int p(\bm{f}_{d}|\bm{u}_{d}, \bX) \log p(y_{n,d}|\bm{f}_{d}, \bm{x}_{n})  d\bm{f}_{d}}_{\mathcal{L}_{f}^{(n,d)}}d\bm{u}_{d}d\bm{x}_{n} \nonumber \\
    &= \sum_{n,d} \int q(\bm{x}_{n}) \underbrace{ \int q(\bm{u}_{d}) \hspace{1mm} \mathcal{L}_{f}^{(n,d)} d\bm{u}_{d}}_{\mathcal{L}_{u}^{(n,d)}}d\bm{x}_{n}\nonumber  \\
    &= \sum_{n,d}  \underbrace{ \int q(\bm{x}_{n})  \hspace{1mm}\mathcal{L}_{u}^{(n,d)} d\bm{x}_{n}}_{\mathcal{L}_{\bX}^{(n,d)}}.   \nonumber
\end{align}

First, performing the integration w.r.t $\bm{f}_{d}$,
\begin{align}
    \mathcal{L}_{f}^{(n,d)} &= \int p(\bm{f}_{d}|\bm{u}_{d}, \bX) \log p(y_{n,d}|\bm{f}_{d}, \bm{x}_{n})  d\bm{f}_{d}\\
     &=  \log \mathcal{N}(y_{n,d}|k^{T}_{n}K_{mm}^{-1}\bm{u}_{d},\sigma^{2}_{y}) -\dfrac{1}{2\sigma^{2}_{y}}q_{n,n}.
\end{align} 

Note: $y_{n,d}$ is a scalar ($d^{th}$ dimension of point $y_{n}$), $k^{T}_{n}$ is a $1\times M$ matrix - the $n^{th}$ row of $K_{nm}$, we know that $p(\bm{f}_{d}|\bm{u}_{d}, \bX) = \mathcal{N}(K_{nm}K^{-1}_{mm}\bm{u}_{d}, K_{nn} - K_{nm}K^{-1}_{mm}K_{mn})$. Further, $\bm{f}_{d}(\bm{x}_{n})$ is a scalar, denoting the value at index $\bm{x}_{n}$ of the vector $\bm{f}_{d}$. $q_{n,n}$ is the $n^{th}$ entry in the diagonal of matrix $Q_{nn} = K_{nn} - K_{nm}K_{mm}^{-1}K_{mn}$

Then, performing the integration w.r.t $\bm{u}_{d}$ (we parameterise $q(\bm{u}_{d}) = \mathcal{N}(\bm{m}_{d}, S_{d})$ as we know its optimal form is a Gaussian and using similar identities as above we),
\begin{equation}
\begin{aligned}
    \mathcal{L}_{u}^{(n,d)}   &=\int q(\bm{u}_{d}) \Big[\log\mathcal{N}(y_{n,d}|k^{T}_{n}K_{mm}^{-1}\bm{u}_{d},  \sigma^{2}_{y}) 
     - \dfrac{1}{2\sigma^{2}_{y}}q_{n,n} \Big] d\bm{u}_{d} \\ \\
 & = \log \mathcal{N}(y_{n,d}|k^{T}_{n}K_{mm}^{-1}\bm{m}_{d}, \sigma^{2}_{y})  -\dfrac{1}{2\sigma^{2}_{y}}q_{n,n} - \dfrac{1}{2\sigma^{2}_{y}}\textrm{Tr}(S_{d}\Lambda_{n}). 
\end{aligned} 
\end{equation}
where $\Lambda_{n} = K_{mm}^{-1}k_{n}k_{n}^{T}K_{mm}^{-1}$ (Note: The $M\times M$ matrix $K_{mn}K_{nm}$ can be decomposed as $\sum\limits_{n=1}^{N}k_{n}k_{n}^{T}$). 
Now, what remains is to perform the integration w.r.t $q(\bm{x}_{n})$. 
\begin{equation}
\begin{split}
    \mathcal{L}_{1} &= \sum_{n,d} \mathcal{L}^{(n,d)}_{\bX} = \sum_{n,d}\log \mathcal{N}(y_{n,d}|\underbrace{\langle k^{T}_{n} \rangle _{q(\bm{x}_{n})}}_{\Psi^{(n,\cdot)}_{1}}K_{mm}^{-1}\bm{m}_{d}, \sigma^{2}_{y})  -\dfrac{1}{2\sigma^{2}_{y}} \textrm{Tr}(\underbrace{\langle K_{nn}\rangle_{q(\bm{x}_{n})}}_{\psi_{0}} - K_{mm}^{-1}\underbrace{\langle K_{mn}K_{nm} \rangle_{q(\bX)}}_{\Psi_{2}}) \\ 
    &- \dfrac{1}{2\sigma^{2}_{y}} \textrm{Tr}(S_{d}K_{mm}^{-1}\underbrace{\langle K_{mn}K_{nm}\rangle_{q(\bm{x}_{n})}}_{\Psi_{2}}K_{mm}^{-1}) \nonumber
    \end{split}
\end{equation}
We note that the only terms involving the latent points $\bm{x}_{n}$ are $K_{nm}$, $K_{nn}$ and $K_{nm}K_{mn}$; due to the summation at the beginning of the equation we can decompose the matrix terms into terms only dependent on the respective data point $\bm{x}_{n}.$

\begin{align}
&= \sum_{n, d} \Big\{ \log \mathcal{N}(y_{n,d}|\underbrace{\langle k(\bm{x}_{n}, Z) \rangle _{q(\bm{x}_{n})}}_{\Psi^{(n,\cdot)}_{1}}K_{mm}^{-1}\bm{m}_{d}, \sigma^{2}_{y}) -\dfrac{1}{2\sigma^{2}_{y}} \textrm{Tr}(\underbrace{\langle k(\bm{x}_{n}, \bm{x}_{n})\rangle_{q(\bm{x}_{n})}}_{\psi_{0}^{n}})  \\ & \hspace{10mm}+ \dfrac{1}{2\sigma^{2}_{y}}\textrm{Tr}(K_{mm}^{-1}\underbrace{\langle k(Z, \bm{x}_{n})k(\bm{x}_{n}, Z)
    \rangle_{q(\bm{x}_{n})}}_{\Psi_{2}^{n}}) 
      - \dfrac{1}{2\sigma^{2}_{y}} \textrm{Tr}(S_{d}K_{mm}^{-1}\underbrace{\langle k(Z, \bm{x}_{n})k(\bm{x}_{n}, Z)
    \rangle_{q(\bm{x}_{n})}}_{\Psi_{2}^{n}}K_{mm}^{-1}) \Big\} \nonumber
\end{align}
where,
\begin{align}
k(\bm{x}_{n},Z) &=  k_{n}^{T} \hspace{2mm}   (\textrm{$n^{th}$ row of matrix $K_{nm}$, dimension $1 \times M$}) \\
k(Z, \bm{x}_{n})k(\bm{x}_{n}, Z) &= k_{n}k^{T}_{n} \hspace{2mm}   (\textrm{dimension $M \times M$})\\
\hspace{2mm}  k(\bm{x}_{n}, \bm{x}_{n}) &= K_{nn}^{(n,n)} \hspace{2mm} (\textrm{$n^{th}$ entry on the diagonal of matrix $K_{nn}$ dimension $1 \times 1$})
\end{align}

\subsection{\texorpdfstring{$\Psi$}{} statistics}
\label{psi_deriv}

In this section we show that expectations of the full covariance matrices  $K_{nm}$, $K_{nn}$ and $K_{nm}K_{mn}$ w.r.t $q(\bX)$ are indeed factorisable  across data points. 

\begin{align}
    \psi_{0} &= \textrm{Tr}( \langle K_{nn}\rangle_{q(\bX)}) = \biggl< \sum_{n=1}^{N} K^{(n,n)}_{nn} \biggr>_{q(\bX)} = \sum_{n=1}^{N}\bigl< K^{(n,n)}_{nn}\bigr>_{q(\bm{x}_{n})}, \hspace{4mm}  (q(\bX) = \prod_{n=1}^{N}q(\bm{x}_{n}))\\
    &= \sum_{n=1}^{N}\psi_{0}^{n}
\end{align}
Next, we look at $\Psi_{1}$,
\begin{align}
    \Psi_{1} = \langle K_{nm}\rangle_{q(\bX)} 
\end{align}

\begin{equation}
    K_{nm} = \begin{bmatrix}
    k(\bm{x}_{1}, \bm{z}_{1}) & \ldots &  k(\bm{x}_{1}, \bm{z}_{M}) \\
    k(\bm{x}_{2}, \bm{z}_{1}) & \ldots &  k(\bm{x}_{2}, \bm{z}_{M}) \\
    \vdots & \vdots & \vdots \\
       k(\bm{x}_{N}, \bm{z}_{1}) &  \ldots & k(\bm{x}_{N}, \bm{z}_{M}) 
    \end{bmatrix} 
    =
    \begin{bmatrix} \longdash & K^{(1,\cdot)}_{nm} & \rlongdash \\
    
    \longdash & K^{(2,\cdot)}_{nm} & \rlongdash \\
     \vdots & \vdots & \vdots \\
     \longdash & K^{(N,\cdot)}_{nm} & \rlongdash \\
        \end{bmatrix}
\end{equation}

\begin{equation}
    \Psi_{1} 
    = \begin{bmatrix}
    \longdash & \Psi^{(1,\cdot)}_{1} & \rlongdash \\
    \longdash & \Psi^{(2,\cdot)}_{1} & \rlongdash \\
     \vdots & \vdots & \vdots \\
    \longdash & \Psi^{(N,\cdot)}_{1} & \rlongdash \\
    \end{bmatrix} =
    \begin{bmatrix} \longdash & \langle K^{(1,\cdot)}_{nm}\rangle_{q(\bm{x}_{1})} & \rlongdash \\
    
    \longdash & \langle K^{(2,\cdot)}_{nm} \rangle_{q(\bm{x}_{2})} & \rlongdash \\
     \vdots & \vdots & \vdots \\
     \longdash & \langle K^{(N,\cdot)}_{nm}\rangle_{q(\bm{x}_{N})} & \rlongdash \\
        \end{bmatrix},
\end{equation}
where we notice that $\Psi_{1}$ is a $N \times M$ matrix where each row just depends on a data point $\bm{x}_{i}$.
\begin{align}
\Psi_{2} &= \langle K_{mn}K_{nm} \rangle_{q(\bX)}\\
&= \begin{bmatrix} \Bigg| & \Bigg| & \vdots & \Bigg|  \\
 \langle K^{(1,\cdot)}_{nm}\rangle_{q(\bm{x}_{1})} & \langle K^{(2,\cdot)}_{nm} \rangle_{q(\bm{x}_{2})} & \ldots & \langle K^{(N,\cdot)}_{nm}\rangle_{q(\bm{x}_{N})} \\
 \Bigg| & \Bigg| & \vdots & \Bigg| \\
\end{bmatrix} 
\begin{bmatrix} \longdash & \langle K^{(1,\cdot)}_{nm}\rangle_{q(\bm{x}_{1})} & \rlongdash \\
    \longdash & \langle K^{(2,\cdot)}_{nm} \rangle_{q(\bm{x}_{2})} & \rlongdash \\
     \vdots & \vdots & \vdots \\
     \longdash & \langle K^{(N,\cdot)}_{nm}\rangle_{q(\bm{x}_{N})} & \rlongdash \\
    \end{bmatrix} \\
&=  \sum_{n=1}^{N}\langle K^{(n,\cdot)^{T}}_{nm} K^{(n,\cdot)}_{nm}\rangle_{q(\bm{x}_{n})}  \\
&= \sum_{n=1}^{N} \Psi_{2}^{n}
\end{align}
which is an $M \times M$ matrix decomposable as a sum of $N$ $M \times M$ matrices where each component matrix is only dependent on a data point $\bm{x}_{i}$.

\subsection{KL divergence between factorised Gaussians}
\label{kl_deriv}
In eq. (5) in the main paper we re-write the KL term involving $q(\bX)$ as a factorisation across $n$, we show the proof below: 
\[
\begin{aligned}
\textrm{KL}(q(\bX)||p(\bX)) &= \textrm{KL}\Big(\prod_{n=1}^{N}q(\bm{x}_{n})||\prod_{n=1}^{N}p(\bm{x}_{n}) \Big) \\
 &= \int\prod_{n=1}^{N}q(\bm{x}_{n})\log \dfrac{\prod_{n=1}^{N}q(\bm{x}_{n})}{\prod_{n=1}^{N}p(\bm{x}_{n})}d\bm{x}_{1}\ldots d\bm{x}_{N}\\
&= \int\prod_{n=1}^{N}q(\bm{x}_{n})\sum_{n=1}^{N}\log \dfrac{q(\bm{x}_{n})}{p(\bm{x}_{n})}d\bm{x}_{1}\ldots d\bm{x}_{N} \\
&= \int\prod_{n=1}^{N-1}q(\bm{x}_{n})q(\bm{x}_{N})\Big ( \log \dfrac{q(\bm{x}_{N})}{p(\bm{x}_{N})} + \sum_{n=1}^{N-1}\log \dfrac{q(\bm{x}_{n})}{p(\bm{x}_{n})}\Big)d\bm{x}_{1}\ldots d\bm{x}_{N} \\
&= \textrm{KL}(q(\bm{x}_{N})||p(\bm{x}_{N}))\underbrace{\int\prod_{n=1}^{N-1}q(\bm{x}_{n})d\bm{x}_{1}\ldots d\bm{x}_{N-1}}_{1} + \textrm{KL}\Big(\prod_{n=1}^{N-1}q(\bm{x}_{n})||\prod_{n=1}^{N-1}p(\bm{x}_{n}) \Big) \\
&=  \sum_{n=1}^{N} \textrm{KL}(q(\bm{x}_{n})||p(\bm{x}_{n}))
\end{aligned}
\]

\section{Further Results}

\subsection{qPCR: Visualisation of Latent Space and relevance parameters}

\begin{figure}
    \centering
    \includegraphics[scale=0.5]{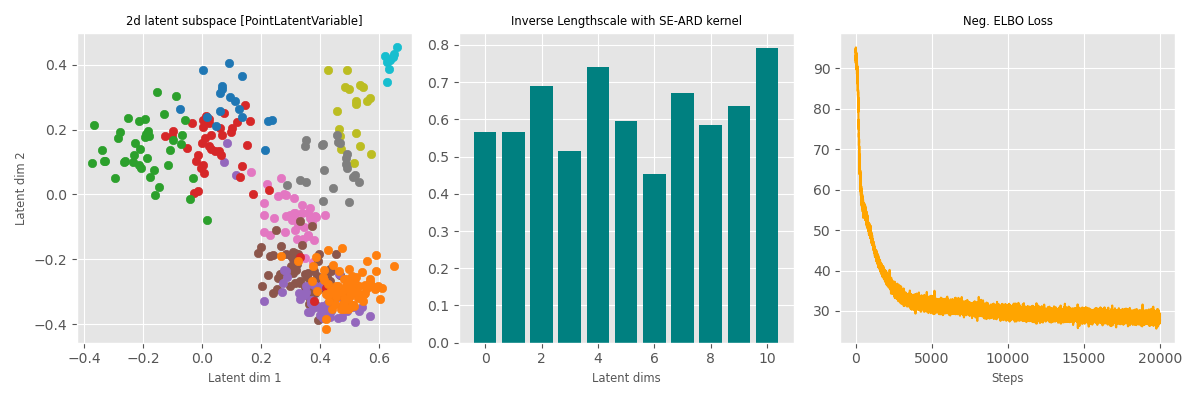}
    \includegraphics[scale=0.5]{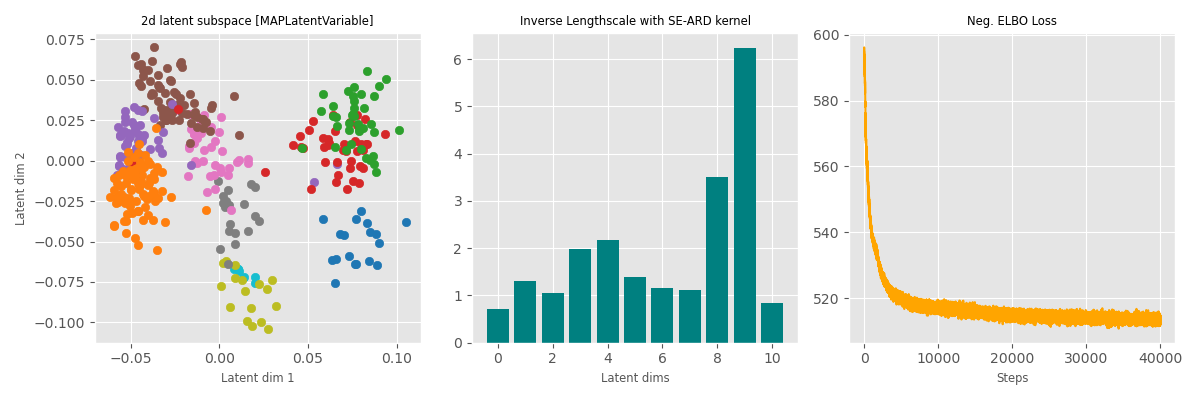}
    \includegraphics[scale=0.5]{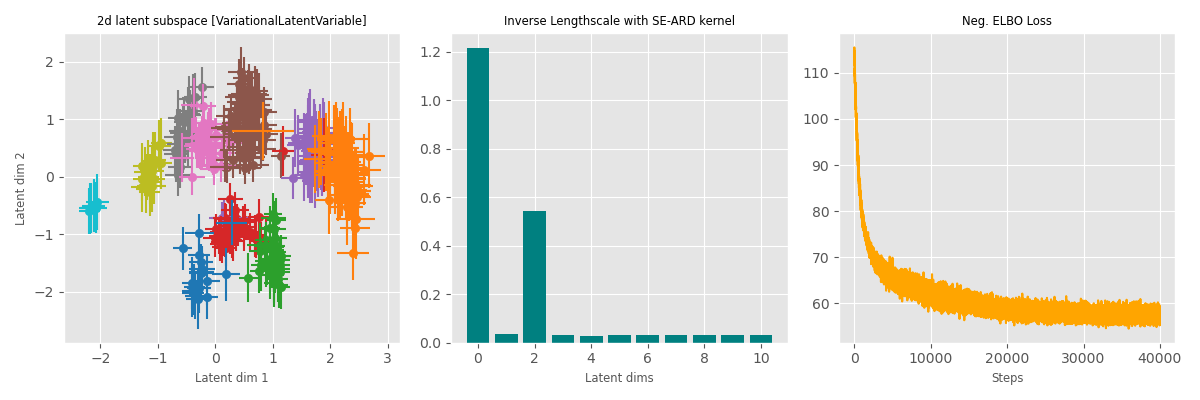}
    \includegraphics[scale=0.5]{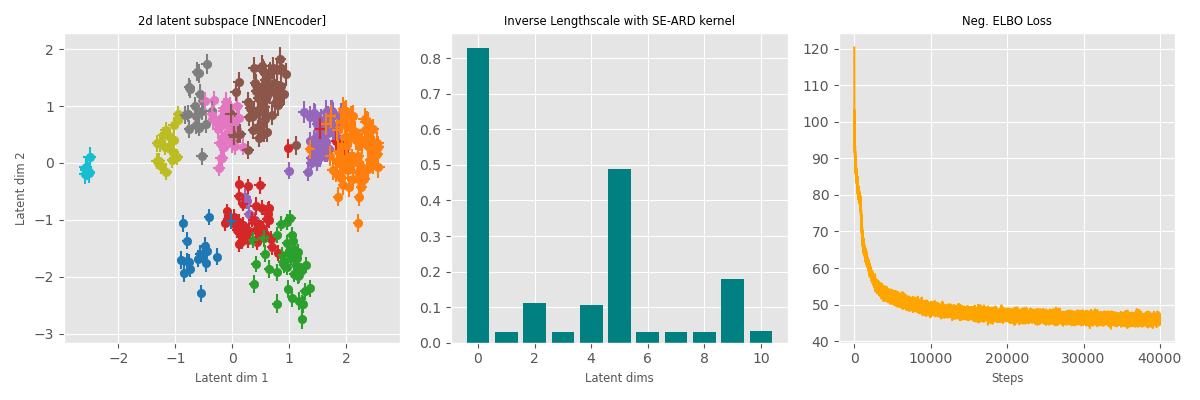}
    \caption{Analysis of qPCR data across models \textsc{Point}, \textsc{MAP}, \textsc{B-SVI} and \textsc{AEB-SVI}. For the bayesian models (bottom 2 rows) the vertical and horizontal lines crossing each point denote axis aligned Gaussian uncertainty of 1 standard deviation.}
    \label{fig:gene}
\end{figure}

We analyse the qPCR dataset and recover the 10 cell developmental stages with our algorithm under each method, however, the point methods underperform the Bayesian methods in terms of disentanglement and also overfit when trained with 11 latent dimensions (see fig. \ref{fig:gene}). Both B-SVI and AEB-SVI give a clean recovery of cell developmental stages from the 48$d$ data. 

\subsection{Oilflow: Automatic Relevance Determination}

\begin{figure}
    \centering
    \includegraphics[scale=0.5]{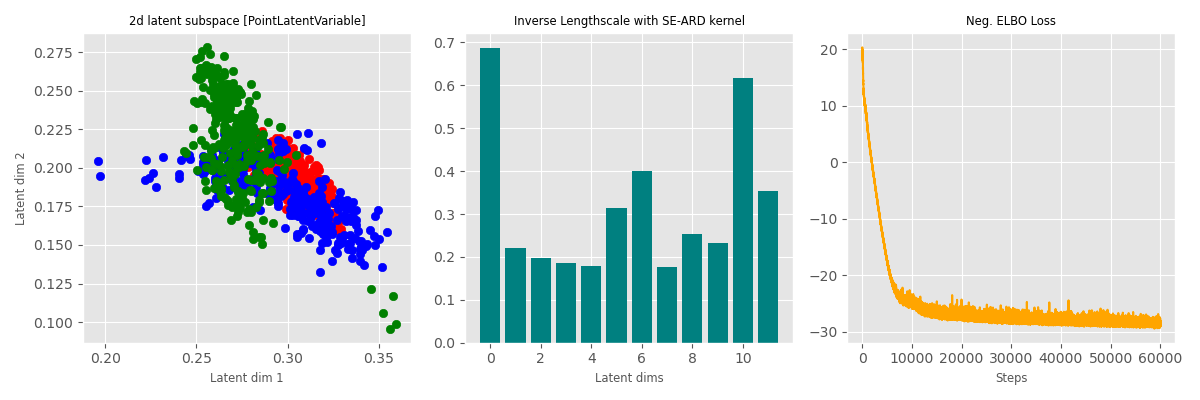}
    \includegraphics[scale=0.5]{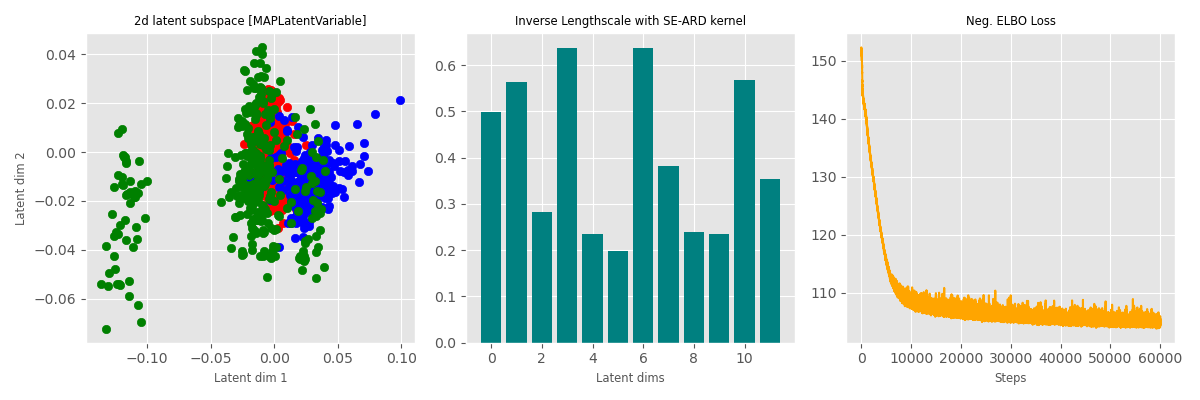}
    \includegraphics[scale=0.5]{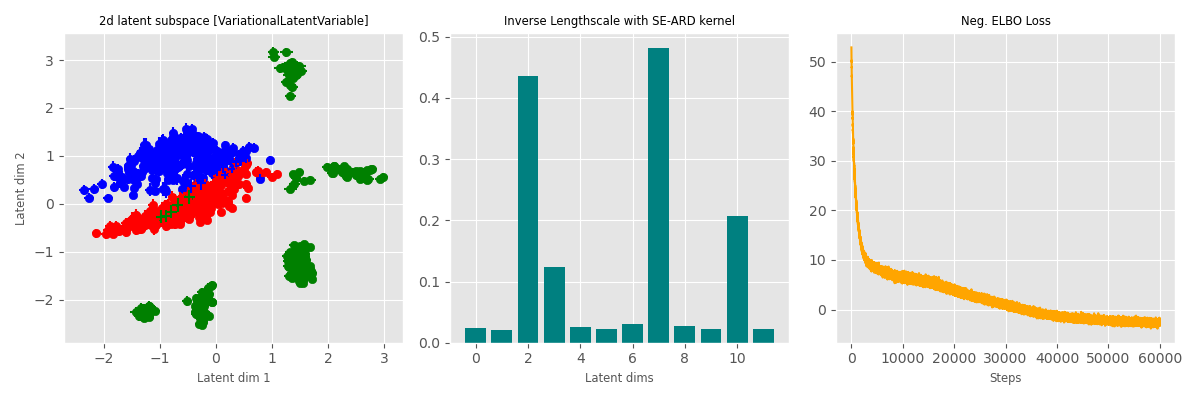}
    \includegraphics[scale=0.5]{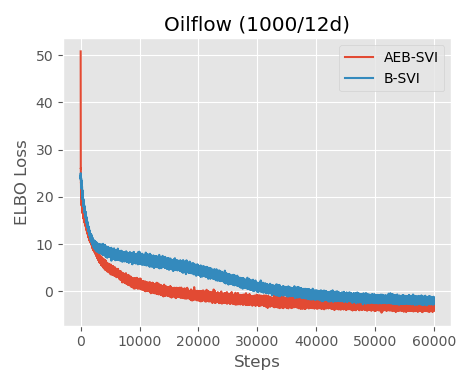}
    \caption{Analysis of oilflow dimensionality reduction with ARD. Final plot shows ELBO loss for B-SVI (non-amortised) and the amortised NNEncoder model with the latter achieving a very similar convergence loss level to the non-amortised model.}
    \label{fig:oilflow2}
\end{figure}

In this experiment we train the oilflow dataset with the same latent dimensions as data dimensions and learn the dominant dimensions from the kernel lengthscales. For the three models \textsc{Point}, \textsc{MAP} and \textsc{B-SVI} the training errors were [0.0074, 0.0105, 0.0590] and test errors were [0.349, 0.527, 0.214] respectively. With more latent dimensions the point methods catastrophically overfit and fail to disentangle the three classes, while B-SVI manages to efficiently recover the dominant dimensions (see fig. \ref{fig:oilflow2}). 

\subsection{NYC Taxi-cab: Test Reconstructions}

\begin{figure}[H]
    \centering
    \includegraphics[scale=0.6]{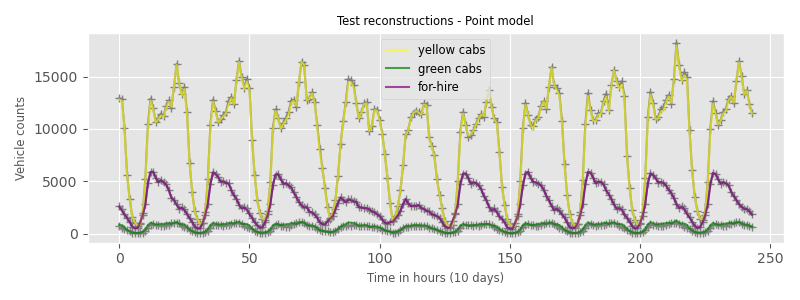}
    \includegraphics[scale=0.6]{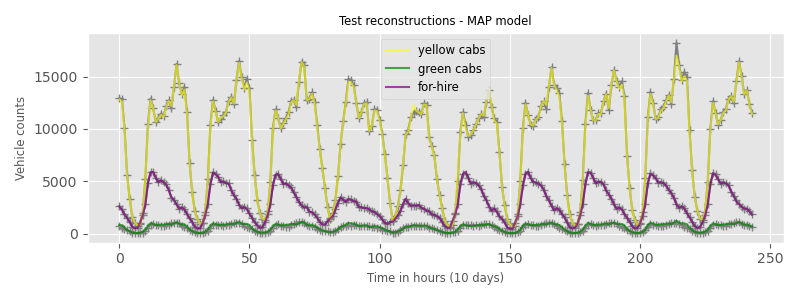}
    \includegraphics[scale=0.6]{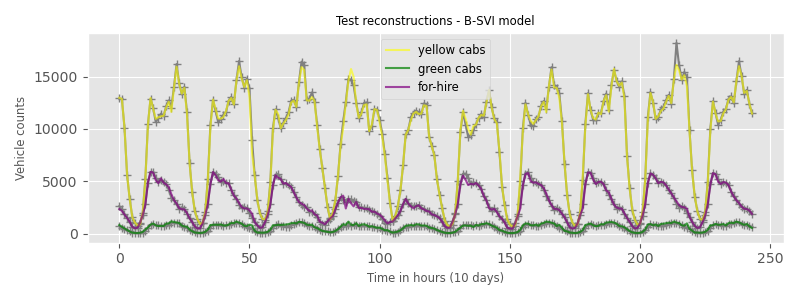}
    \includegraphics[scale=0.6]{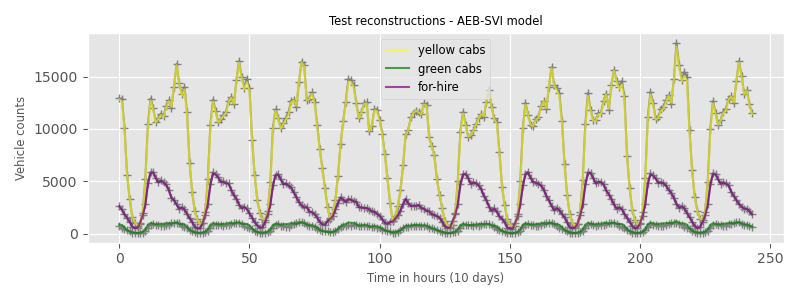}
    \caption{Test reconstructions per taxi type over a 10 day period. All models are able to reconstruct the test time vehicle counts accurately, the spike on day 9 for yellow cabs is marginally underestimated for the Gaussian (B-SVI) and MAP model for this run, the error from that spike contributes $\approx 122$ to the average RMSE. The test RMSE's for the above plots are: 97.12, 104.97, 176.51, 97.03 for the 4 models (in order of above) respectively.}
    \label{fig:taxi}
\end{figure}

In the plots in fig. \ref{fig:taxi} we visualise the ground-truth and predicted reconstructions per dimension (this corresponds to the 3 different taxi types operating in NYC namely - yellow, green and for-hire) over the whole test period of 10 days, each point indicates the total number of trips per hour.

\clearpage
\subsection{MNIST: Missing data reconstructions}
The plots below demonstrate reconstruction abilities when the algorithm is trained only on partially observed data. The missing pixels are distributed randomly for every image. Despite masking a large fraction of the pixels, the correct structure is reconstructed with only marginal degradation. 
\begin{figure}[H]
\includegraphics[scale=0.41]{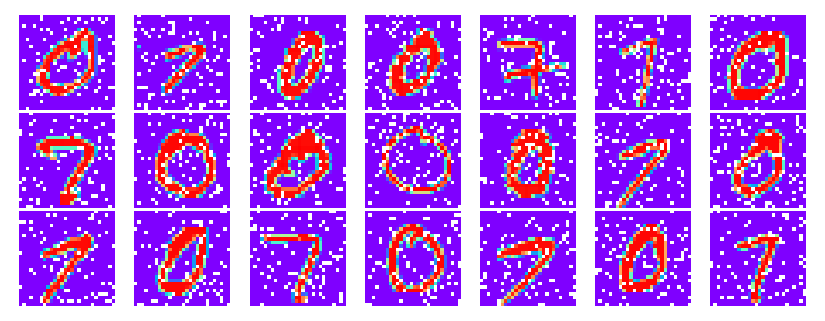}
    \includegraphics[scale=0.43]{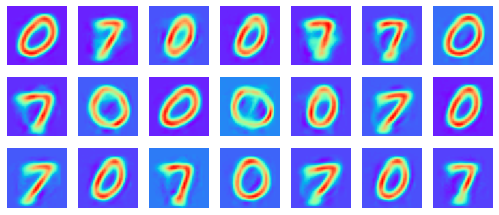} \\
    \includegraphics[scale=0.38]{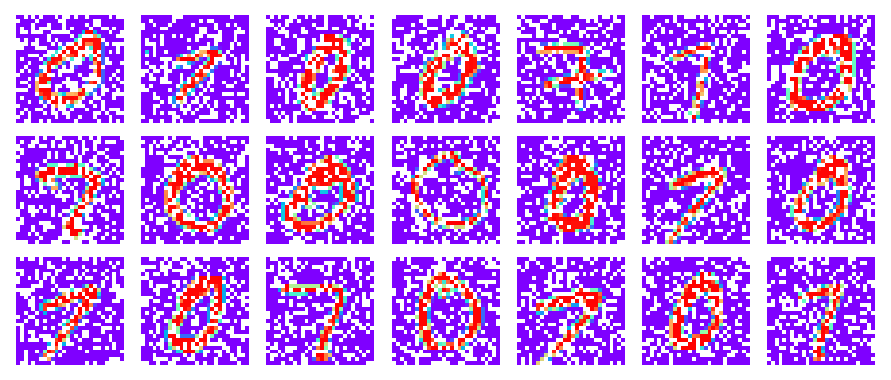}
    \includegraphics[scale=0.5]{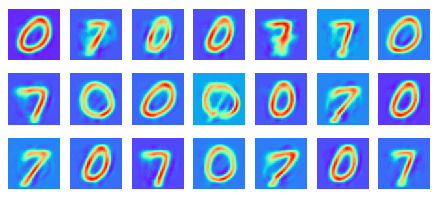} \\
    \includegraphics[scale=0.4]{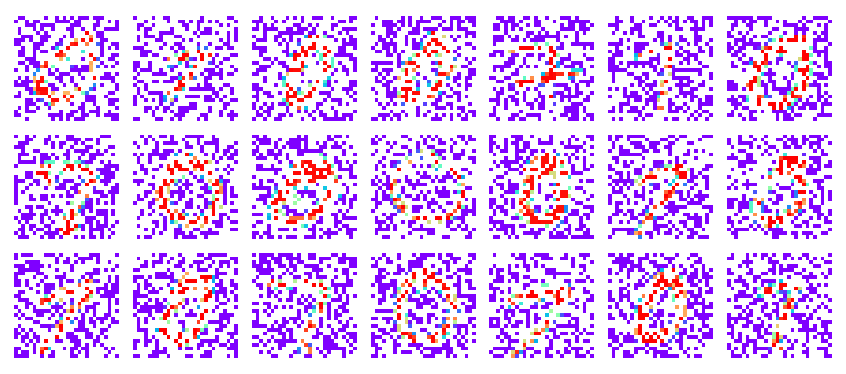}
    \includegraphics[scale=0.48]{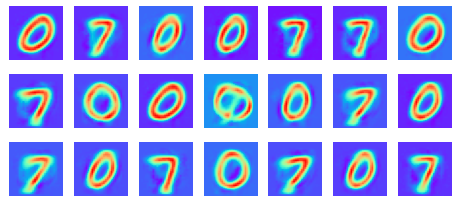}
    \caption{MNIST images reconstruction task for different degrees of missingness (10$\%$, $30\%$, $60\%$). Left: Training digit. Right: Reconstruction post training.}
    \label{fig:mnist}
\end{figure}

\subsection{MOCAP: Missing data reconstructions}
We train \textsc{B-SVI} on 62$d$ human motion capture data to try and recreate the sequence of diverse motions (walking, jumping and running) for a single subject. In order to test the models ability to learn in the presence of missing data we remove dimensions corresponding to different body parts to simulate different types of \textit{missingness}. We cycle over the following types of structural missingness: (a) missing head, right leg and forearm, (b) missing forearms and left leg (c) missing upper body (d) missing lower body. Overall, the model yields very sensible reconstructions given the challenge of arbitrarily missing data. We note that the walking motion has the best reconstruction while jumping and running yield much superior reconstructions at training rather than test time. This is because the model had never seen the arm motion during running (arm data was missing from the training point) hence at test time the arm motion defaults to walking but the leg strides are captured accurately. 

\begin{figure}[H]
    \centering
    \includegraphics[scale=0.79]{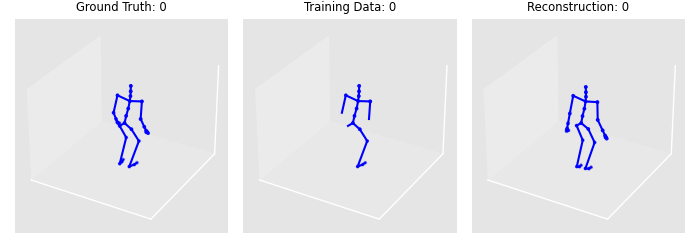}
    \includegraphics[scale=0.79]{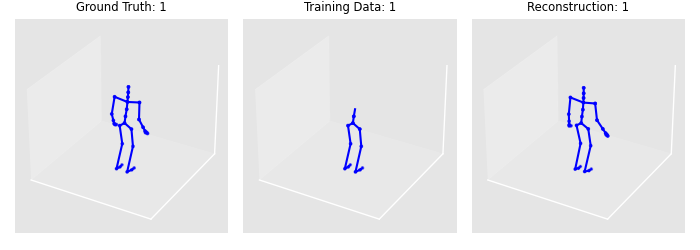}
    \includegraphics[scale=0.79]{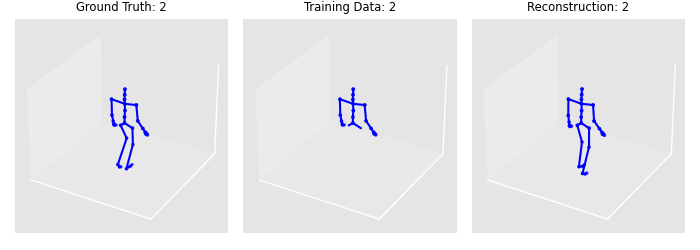}
    \includegraphics[scale=0.79]{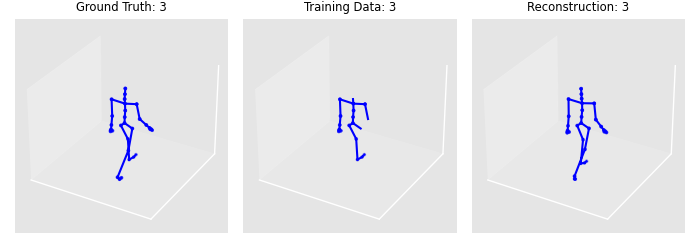}
    \caption{MOCAP reconstructions of missing dimensions on training data}
    \label{fig:mocap_train}
\end{figure}\clearpage
\begin{figure}[H]
    \centering
    \includegraphics[scale=0.79]{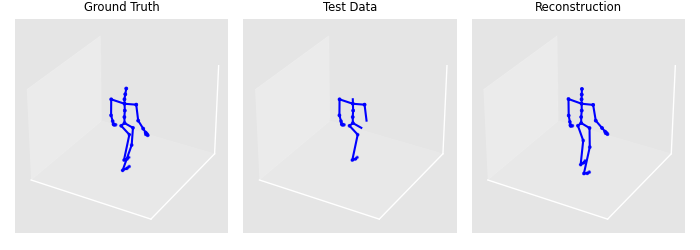}
    \includegraphics[scale=0.79]{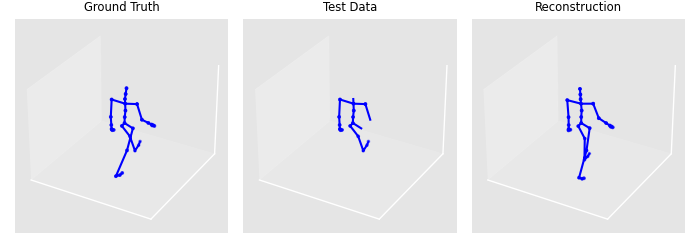}
    \includegraphics[scale=0.79]{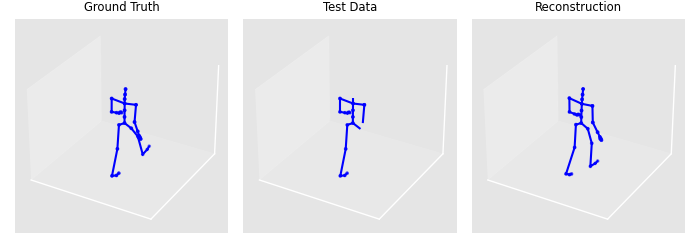}
    \includegraphics[scale=0.79]{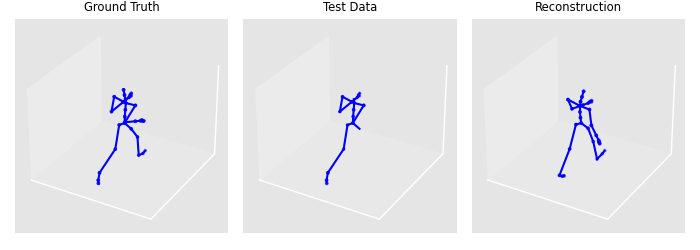}
    \caption{MOCAP reconstructions of missing dimensions on test data}
    \label{fig:mocap_test}
\end{figure}

\subsection{Flexible Variational Families using Normalising Flows}

In this section, we briefly introduce the use of normalising flows in our framework for capturing richer, non-Gaussian distributions in latent space.

Instead of parameterising $q(\bX)$ as a Gaussian we can parameterise it as a transformed Gaussian distribution by using a sequence of invertible and differentiable transformations. The variational distribution of \textsc{B-SVI} with a transformed Gaussian distribution is given by, 
\begin{equation*}
q(\bX) = \prod_{n=1}^{N}\mathcal{N}(\bm{x}_{n}; \bm{\mu}_{n}, s_{n}\mathbb{I}_{Q}) \Bigg\lvert\det \prod_{j=1}^{k} \dfrac{\partial g_{j}}{\partial \bm{x}_{n}^{(j-1)}} \Bigg \rvert^{-1}, \\ 
\label{mf_flow}
\end{equation*}
where the parameters of the flow mappings $g_{j}$ are collected in $\zeta$. We perform Monte Carlo expectations of the terms in the uncollapsed lower bound that involve $q_{\bm{\phi}}(\bm{x}_{n})$ by sampling from the base Gaussian at each step $\bm{x}_{n}^{(0)} \sim \mathcal{N}(\mu_{n}, s_{n}\mathbb{I}_{Q})$ and passing them through the flow $g_{k} \circ g_{k-1} \circ \ldots g_{1} (\bm{x}_{n}^{(0)})$ to yield the final latent point $\bm{x}_{n}^{(k)}$. 

One can model each row of $\bX$ (latent point) by an independent flow, although the number of parameters to be estimated in this model becomes too unwieldy very quickly. This approach also does not address the independence assumption across different data points. An alternative model that shares flow parameters across different latent points is tractable but is highly constrained to learn a flexible distribution per individual latent point and the trained flow-based distributions appear to be rather similar to a multivariate normal latent distribution only capturing local correlations but not non-linear correlations.

\begin{figure}[h]
    \centering
    \includegraphics[scale=0.3]{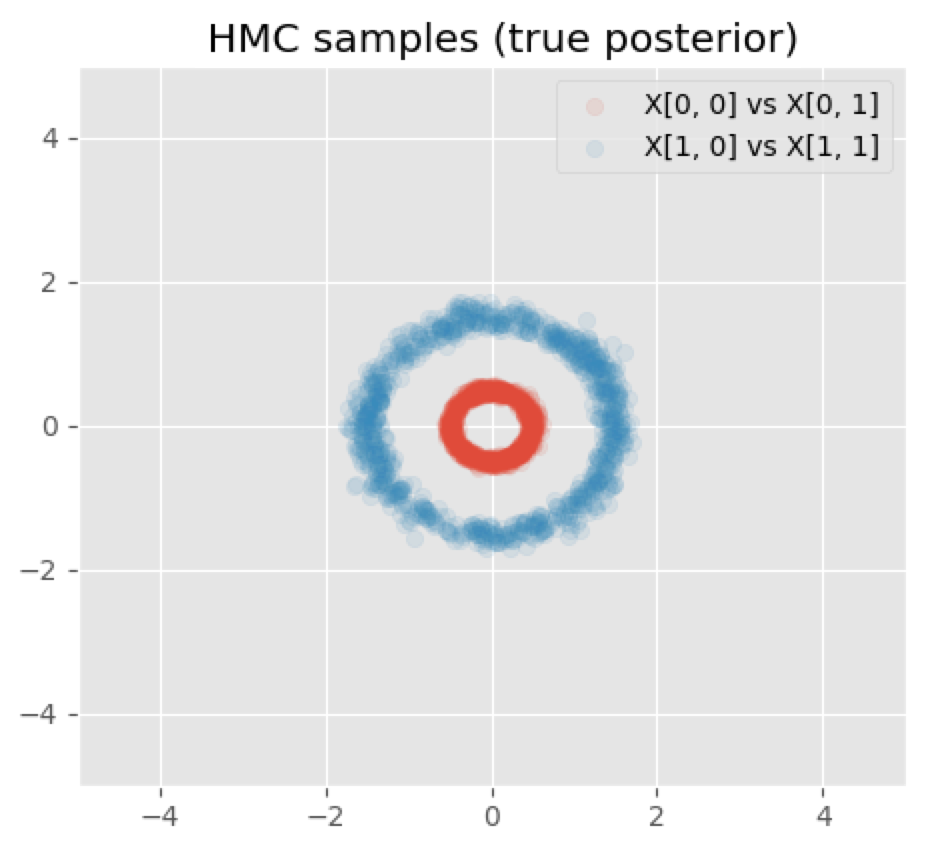}
    \includegraphics[scale=0.4]{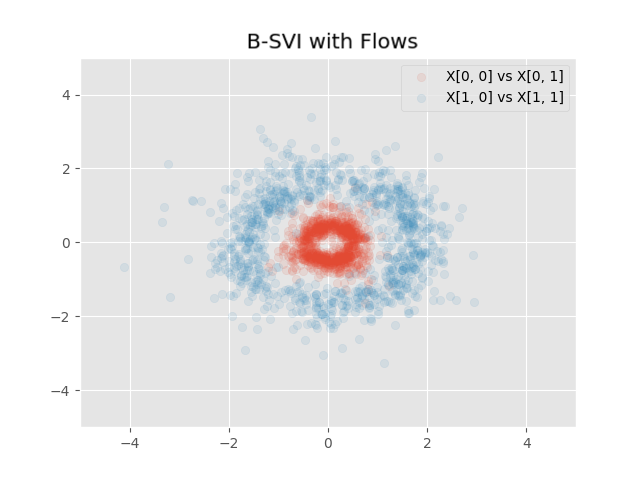}
    \caption{A demonstration of the effectiveness of the doubly stochastic algorithm to learn normalising flow based latent distributions for two individual high-dimensional points. \textit{Left:} HMC samples from the true posterior. \textit{Right:} Samples from the optimised flow based variational distribution with a base Gaussian and a sequence of 40 planar flows.}
    \label{fig:hmc}
\end{figure}

An interesting case is to use a single flow to model the joint density $q(\bX)$, so in the case of 2 latent points in 2$d$ we learn a four dimensional flow based distribution modelling vec$(\bX)$. Unsurprisingly, the distributions learnt in this model closely approximate the true posterior of the latent variables obtained using HMC (see fig. \ref{fig:hmc}).

In this demo experiment, we generate a toy synthetic dataset using the forward model of the GPLVM (with a linear kernel, so this is equivalent to probabilistic PCA) of two points in $\mathbb R^{10}$, we attempt to learn a 2d latent space for this toy 2 point dataset using Algorithm 1. described in the paper except that $q(\bX)$ is non-Gaussian. We visualise samples from posterior distributions corresponding to the two 10$d$ points in latent space. 

As PCA is rotation invariant, the true posterior of the latent variable is unidentifiable under rotations. This can be seen in the HMC samples drawn from the posterior of each latent point. We see that the SVI based model augmented with planar flows is able to capture this non-Gaussianity. 

However, to achieve this result, many optimisation tricks were employed (e.g. increasing the number of samples to approximate the KL-divergence calculation, using a very small learning rate 1e-05 and long training times). The quality of results for flow based variational families for small toy-examples closely resemble gold-standard HMC but optimisation remains a difficulty and further research is required to understand how to achieve good training performance for moderate sized datasets.

\section{Experimental Configuration}
\vspace{-3mm}
\begin{table}[H]
    \centering
    \begin{tabular}{c|c|c|c|c|c|c|c}
        Dataset & $N$ & $D$ & $Z$ & $Q$ & LR  & Mini-batch & Train w. missing \\
        \hline 
         Oilflow & 1000 & 12 & 25 & 10 & 1e-03  & 100 & No\\
         qpCR &  450 & 48 & 40 & 11 & 1e-03  & 100 & No \\
         Taxi-cab & 744 & 3 & 36 & 2 & 5e-03  & 500 & No \\
         MNIST & 15K & 768 & 100 & 5 & 0.01  & 100 & Yes\\
         Brendan & 1965 & 560 & 120 & 5 & 0.01  &  450 & Yes \\
         MOCAP & 533 & 62 & 30 & 6 & 0.01  & 200 & Yes\\
         MovieLens & 943 & 1682 & 34 & 15 & 0.005  & 100 & Yes \\
    \end{tabular}
    \caption{Training experimental configuration where $N$ and $D$ denote the number of data points and data space dimensions, $Z$ denotes the number of inducing inputs shared across dimensions, $Q$ denotes the dimesionality of the latent space, LR denotes the learning rate, $\beta$ denotes the scalar annealing factor for the KL latent term in the ELBO.}
    \label{tab:exp}
\end{table}

\section{AEB-SVI: Network Architecture} 

We use two separate MLPs to encode the mean and covariance matrix, we use 2 hidden weight layers with tanh non-linearity. We summarise the network architecture (with the input and output layers) for learning the mean vector and covariance matrix below:

-Oilflow $(12D)$ 

a. Mean network: $(12 (D), 10, 5, 12 (Q))$
b. Covariance network: $(12 (D), 78, 78, 144 (Q^{2}))$

The number of nodes in the hidden layers for the covariance network are derived as $(D + Q^{2})/2$ 

-qPCR $(48D)$

a. Mean network: $(48 (D), 10, 5, 11 (Q))$
b. Covariance network: $(48 (D), 84, 84, 121 (Q^{2}))$

-Taxi-cab $(3D)$

a. Mean network: $(3 (D), 3, 3, 2 (Q))$
b. Covariance network: $(3 (D), 3, 3, 4 (Q^{2}))$

Overall, we found that the latent representations learnt in the amortised case where not extremely sensitive to the architecture as long as sufficient capacity was reached. 
For qPCR, the latent dimensionality of $Q=11$ was high enough to express the effective dimensionality of the data as the test reconstructions did not improve much for a higher latent dimensionality. Although, selecting $Q=48$ for fully automatic model selection would not overfit in the Bayesian case, it would increase the compute time due the number of variational parameters and it is sensible to fine-tune $Q$ to an appropriate size. $Q$ was fixed across models to facilitate a comparison. 

\section{Code Contribution}

We wrote a custom implementation of Bayesian GPLVM in the \texttt{gpytorch} library  \citep{gardner2018gpytorch} to run the various latent variable configurations in this paper. We also wrote a custom Gaussian missing data likelihood class that can seamlessly handle NaNs in the $\bY$ data matrix. The code can be found attached to supplementary material and is publicly available.

\end{document}